\documentclass[11pt,twoside]{article}

\usepackage[T1]{fontenc} 
\usepackage{microtype} 
\usepackage[hmarginratio=1:1,outer=28mm,top=23mm,bottom=28mm]{geometry} 
\usepackage{times} 
\usepackage{graphicx} 
\usepackage{natbib} 
\usepackage{algorithm} 
\usepackage{algorithmic}
\usepackage{color}
\usepackage[usenames,dvipsnames]{xcolor}
\usepackage[ocgcolorlinks,pdfpagemode={UseOutlines},bookmarks=true,bookmarksopen=true, bookmarksopenlevel=0,bookmarksnumbered=true,hypertexnames=false, hidelinks,   pdfstartview={FitV},unicode,breaklinks=true]{hyperref} 

\usepackage{appendix}
\usepackage{chngcntr} 
\usepackage[super]{nth} 
\usepackage{adjustbox} 
\usepackage{capt-of}
\usepackage{subcaption}
\usepackage{sidecap}
\usepackage{caption}
\usepackage{float}
\usepackage{xfrac}
\usepackage{todonotes}
\usepackage{siunitx}
\usepackage{enumitem}
\usepackage[eulergreek]{sansmath} 
\usepackage{tikz}
\usetikzlibrary{decorations.pathreplacing}
\usetikzlibrary{decorations.markings}
\usetikzlibrary{patterns}
\usetikzlibrary{positioning}
\usetikzlibrary{shapes,arrows}
\usetikzlibrary{calc}
\usetikzlibrary{intersections}
\usepackage{pgfplots}
\usetikzlibrary{pgfplots.groupplots}
\usepackage{amssymb} 
\usepackage{amsmath}
\usepackage{booktabs}
\usepackage{cases}
\usepackage{cleveref}
\usepackage{tabularx}
\usepackage{wrapfig}

\usepackage{abstract} 

\usepackage{titlesec} 

\usepackage{fancyhdr} 
\crefname{equation}{eq.}{eqs.}
\crefname{table}{tab.}{tabs.}
\crefname{figure}{fig.}{figs.}
\crefname{section}{sec.}{secs.}
\crefname{pluralequation}{eqs.}{eqs.}
\Crefname{pluralequation}{Eqs.}{Eqs.}
\crefformat{equation}{#2eq.~(#1)#3}
\Crefformat{equation}{#2Eq.~(#1)#3}
\crefformat{pluralequation}{#2eqs.~(#1)#3}
\Crefformat{pluralequation}{#2Eqs. (#1)#3}
\crefrangeformat{pluralequation}{eqs.~(#3#1#4) to~(#5#2#6)}
\Crefrangeformat{pluralequation}{eqs.~(#3#1#4) to~(#5#2#6)}
\crefmultiformat{pluralequation}{eqs.~(#2#1#3)}{ and~(#2#1#3)}{, (#2#1#3)}{ and~(#2#1#3)}
\Crefmultiformat{pluralequation}{Eqs.~(#2#1#3)}{ and~(#2#1#3)}{, (#2#1#3)}{ and~(#2#1#3)}
\allowdisplaybreaks
\graphicspath{{../figs/}}
\definecolor{Red}{rgb}{1.0,0,0} 
\definecolor{Green}{rgb}{0,.80,0} 
\definecolor{Blue}{rgb}{0,0,0.80} 

\newcommand{\RRR}{\mathbb{R}}
\newcommand{\classk}{k}
\newcommand{\classK}{K}
\newcommand{\subc}{c}
\newcommand{\subC}{C}
\newcommand{\dimd}{d}
\newcommand{\dimD}{D}
\newcommand{\inpn}{n}
\newcommand{\inpN}{N}
\NewDocumentCommand{\y}{O{}O{}}{y_{#2}^{\,#1}\!}
\NewDocumentCommand{\yVec}{O{}}{\vec{\y}^{\,#1}\!}
\newcommand{\yVecN}{\yVec[(\inpn)]}
\newcommand{\yN}{\y[\!(\inpn)]}
\newcommand{\ydN}{\y[\!(\inpn)][\dimd]}
\newcommand{\yd}{\y[][\dimd]\,}
\newcommand{\labely}{l}
\newcommand{\labelyN}{l^{(n)}\!}
\NewDocumentCommand{\uu}{O{}O{}}{u_{#1}^{\,#2}\!}
\newcommand{\uk}{\uu[\classk]}
\newcommand{\ukN}{\uu[\classk][{(n)}]\!}
\newcommand{\uVec}{\vec{\uu}}
\newcommand{\uVecN}{\vec{\uu}^{\,(n)}\!}
\NewDocumentCommand{\Wgen}{O{}O{}}{\mathcal{W}_{#1#2}}
\newcommand{\Wgencd}{\Wgen[\subc][\dimd]}
\NewDocumentCommand{\Rgen}{O{}O{}}{\mathcal{R}_{#1#2}}
\newcommand{\Rgenkc}{\Rgen[\classk][\subc]}
\NewDocumentCommand{\W}{O{}O{}}{W_{\!#1#2}}
\newcommand{\Wcd}{\W[\subc][\dimd]}
\newcommand{\WcdN}{\W[\subc][\dimd]^{(\inpn)}}
\NewDocumentCommand{\R}{O{}O{}}{R_{#1#2}}
\newcommand{\Rkc}{\R[\classk][\subc]}
\newcommand{\RkcN}{\R[\classk][\subc]^{(\inpn)}}
\NewDocumentCommand{\Sc}{O{\subc}O{}}{s_{#1}^{#2}}
\newcommand{\ScN}{\Sc[\subc][(\inpn)]\!}
\NewDocumentCommand{\sVec}{O{}}{\vec{s}_{\vphantom{\subc}}^{\,#1}}
\newcommand{\sVecN}{\sVec[(\inpn)]\!}
\NewDocumentCommand{\Igenc}{O{\subc}O{}}{\mathcal{I}_{#1}^{#2}}

\NewDocumentCommand{\Ic}{O{\subc}O{}}{I_{#1}^{#2}}
\newcommand{\IcN}{\Ic[\subc][(\inpn)]\!}
\NewDocumentCommand{\tk}{O{\classk}O{}}{t_{#1}^{#2}}
\newcommand{\tkN}{\tk[\classk][(\inpn)]\!}
\NewDocumentCommand{\tVec}{O{}}{\vec{t}_{\vphantom{\classk}}^{\,#1}}

\NewDocumentCommand{\eps}{O{}}{\epsilon_{\textnormal{\tiny $#1$}}}
\newcommand{\eW}{\eps[\W]}
\newcommand{\eR}{\eps[\R]}
\NewDocumentCommand{\epst}{O{}}{\tilde{\epsilon}_{\textnormal{\tiny $#1$}}}

\newcommand{\normA}{A}

\newcommand{\BvSB}{\mathrm{BvSB}}

\newcommand{\ThetaOld}{\Theta^{\mathrm{old}}}
\newcommand{\FF}{{\cal F}}
\newcommand{\E}[1]{\big\langle{}#1\big\rangle}

\newcommand{\Wnn}{W}
\newcommand{\Wnncd}{\Wnn_{cd}}
\newcommand{\epss}{\epsilon}
\newcommand{\dPrime}{d^{\prime}}
\newcommand{\cPrime}{c^{\prime}}
\newcommand{\nPrime}{n^{\prime}}

\newcommand{\invisible}[1]{}   

\title{\vspace{-10pt}\LARGE Neural Simpletrons -- Minimalistic Directed Generative Networks for Learning with Few Labels}

\author{
\large
Dennis Forster$^{1,3}$, Abdul-Saboor Sheikh$^{2,3}$, J\"org L\"ucke$^1$ \\
\normalsize dennis.forster@uni-oldenburg.de, sheikh.abdulsaboor@gmail.com, joerg.luecke@uni-oldenburg.de\\[2mm]
\normalsize $^1$\,Carl von Ossietzky University of Oldenburg, Oldenburg, Germany \\
\normalsize $^2$\,Technical University of Berlin, Berlin, Germany \\ 
\normalsize $^3$\,Frankfurt Institute for Advanced Studies (FIAS), Frankfurt am Main, Germany \\[2mm]
}
\date{}

\begin{document}

\maketitle

\begin{abstract}
Classifiers for the semi-supervised setting often combine strong supervised models with additional learning objectives to make use of unlabeled data.
This results in powerful though very complex models that are hard to train and that demand additional labels for optimal parameter tuning, which are often not given when labeled data is very sparse.
We here study a minimalistic multi-layer generative neural network for semi-supervised learning in a form and setting as similar to standard discriminative networks as possible. 
Based on normalized Poisson mixtures, we derive compact and local learning and neural activation rules.
Learning and inference in the network can be scaled using standard deep learning tools for parallelized GPU implementation.
With the single objective of likelihood optimization, both labeled and unlabeled data are naturally incorporated into learning.
Empirical evaluations on standard benchmarks show, that for datasets with few labels the derived minimalistic network improves on all classical deep learning approaches and is competitive with their recent variants without the need of additional labels for parameter tuning.
Furthermore, we find that the studied network is the best performing monolithic (`non-hybrid') system for few labels, and that it can be applied in the limit of very few labels,
where no other system has been reported to operate so far.
\end{abstract}
\vspace{5mm}
\section{Introduction}
Deep neural networks (DNNs) have demonstrated state-of-the-art performance in many application domains.
If large labeled databases and large computational resources are available, discriminative deep networks are now among the best performing systems in tasks such as image or speech recognition, document classification and many more \citep[for example,][]{Schmidhuber2015,BengioEtAl2013,HintonEtAl2012}.

If no labels are available, unsupervised approaches are the method of choice, and those based on deep directed graphical models are well suited to capture the rich structure of typical data such as images or speech.
However, while being potentially more powerful information processors than discriminative systems, such directed models are typically trained on much smaller scales (either because of computational limits or performance saturation).
For instance, deep sigmoid belief networks \citep[SBNs,][]{SaulEtAl1996,GanEtAl2015} or newer models such as NADE \citep[][]{LarochelleMurray2011} have only been trained with a couple of hundred to about a thousand hidden units \citep[][]{BornscheinBengio2015,GanEtAl2015}.

For settings of partly labeled training data, supervised and unsupervised approaches come together.
These semi-supervised settings are increasingly interesting both for technical and practical reasons:
While obtaining large amounts of data (like images or sounds) is often relatively easy, the effort to obtain labels is comparably high, for example, if manual hand-labeling of the data is required.
Data sets with few labels therefore emerge as a natural application domain, and such settings have consequently shifted into the focus of many recent contributions \citep[][]{LiuEtAl2010,WestonEtAl2012,PitelisEtAl2014,KingmaEtAl2014,RasmusEtAl2015,MiyatoEtAl2015}.

The most successful contributions in this semi-supervised setting so far have been hybrid combinations of two or more learning algorithms, which
merge unsupervised and supervised learning \citep[see, for example,][]{WestonEtAl2012,KingmaEtAl2014,RasmusEtAl2015,MiyatoEtAl2015}.
However, while deep neural networks alone are often already equipped with many tunable parameters (for architecture, regularization, sparsity etc.), such hybrid approaches add further parameters for the interplay between supervised and unsupervised learning.
This makes their practical application to settings where only few labels are available difficult: In principle those models are able to train on very small amounts of labeled data with state-of-the-art results.
However, to find suitable settings of tunable parameters for such complex models, generally many more labels are needed than available during training in order to avoid the risk of highly overfitting to a very small validation set.
Consequently, similar performance has never been shown when not only the amount of training labels, but the total amount of labels (that is, during training and tuning) was highly restricted.

This work investigates the semi-supervised setting with a minimalistic, deep directed graphical model, which can be formulated as a neural network.
The objective of likelihood optimization given by the graphical model directly combines information of unlabeled and labeled data in a monolithic learning system.
With only a handful of resulting free parameters, tuning can be done even in settings, where labeled data is extremely sparse.
Furthermore, the similarity to standard neural networks enables the application of software tools for parallelized learning on GPUs \citep[like][]{BastienEtAl2012}.
This allows to scale the generative network to (ten-)thousands of hidden units (we here show networks with up to \num{20000} hidden units) and to apply it to large data sets (here, up to $\approx$\,\num{400000} samples).
Finally, the use of local and compact inference and learning rules closely links the network to recent approaches for bio-inspired computer hardware, such as VLSI \citep[especially][]{NeftciEtAl2015,DiehlCook2015,NesslerEtAl2013}.

\section[A Hierarchical Mixture Model]{A Hierarchical Mixture Model for Classification}
\label{SecGenerativeModel}
A classification problem can be modeled as an inference task based on a probabilistic mixture model.
Such a model can be hierarchical, or {\em deep}, if we expect the data to obey a hierarchical structure.
For hand-written digits, for instance, we first assume the data to be divided into digit classes (`\num{0}' to `\num{9}') and within each class, we expect structure that distinguishes between different writing styles.
Most deep systems allow for a much deeper substructure, using five, ten, or recently even up to 100 or 1000 layers \citep{HeEtAl2015}.
For our goal of semi-supervised learning with few labels, we however want to restrain the model complexity to the necessary minimum of a hierarchical model.

\subsection{The Generative Model} 
\begin{wrapfigure}[11]{r}{0.215\textwidth}
  \captionsetup{format=plain}
	\centering
	\resizebox{0.175\textwidth}{!}{
		\begin{adjustbox}{trim=0pt 0pt 10pt 8pt}%
			\begin{tikzpicture}[shorten >=1pt,shorten <=1pt,draw=black!50,inner sep=0.333em, outer sep=0.5\pgflinewidth]
	\tikzstyle{neuron}=[circle,fill=black!25,minimum size=17pt,inner sep=0pt]
	\tikzstyle{input neuron}=[neuron, draw = black!50, fill = gray!50]; 
	\tikzstyle{subc neuron}=[neuron, draw = black!50, fill = gray!25];
	\tikzstyle{classk neuron}=[neuron, draw = black!50, fill = gray!5];
	\def\layerdistance{1.75cm}
	\def\nodedistance{0.9cm}
	
	\node[input neuron] (I) at (\nodedistance,0) {$\yVec$};
	
	\node[subc neuron] (C) at (\nodedistance, \layerdistance) {$\subc$};

	\node[classk neuron] (K) at (\nodedistance, 2*\layerdistance) {$\classk$};
	
	\node[input neuron] (L) at (0,0) {$\labely$};

	\path (K) edge [->] node [right, pos=0.5] {$\Rgenkc$} (C);
	\path (C) edge [->] node [right, pos=0.5] {$\Wgencd$}(I);
	\path (K) edge [-, shorten <= 0pt, shorten >= 0pt] (0, 2*\layerdistance);
	\path (0, 2*\layerdistance) edge [->, shorten <= 0pt] (L);
\end{tikzpicture}
		\end{adjustbox}
	}
	\caption{Graphical illustration of the hierarchical generative model.}
	\label{fig:generativemodel}
\end{wrapfigure}
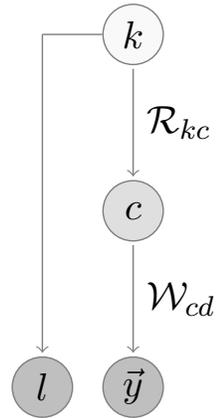
In accordance with the hierarchical formulation of a classification problem, we define the minimalistic hierarchical generative model shown in \cref{fig:generativemodel} as follows:
\begin{align}
&p(\classk) = \frac{1}{\classK}, \qquad p(\labely|\classk) = \delta_{\labely\classk} \label{eq:PriorK}\\[6pt]
&p(\subc|\classk,\Rgen) = \Rgenkc, &&\sum_\subc \Rgenkc = 1 \label{eq:PriorC}\\
&p(\yVec\,|\subc,\Wgen) = \prod_\dimd \mathrm{Poisson}(\yd;\Wgencd), &&\sum_\dimd \Wgencd = \normA\label{eq:Obs} 
\end{align}

The parameters of the model, $\Wgen\!\in\!\RRR_{> 0}^{\subC\times\dimD}$ and $\Rgen\!\in\!\RRR_{\ge 0}^{\classK\times\subC}$, will be referred to as generative weights, which are normalized to constants $\normA$ and $1$, respectively.
The top node (see \cref{fig:generativemodel}) represents $\classK$ abstract concepts or super classes~$\classk$ with labels~$\labely$ (for example, digits `\num{0}' to `\num{9}').
The middle node represents any of the occurring $\subC$ subclasses~$\subc$ (like different writing styles of the digits).
And the bottom nodes represent an observed data sample $\yVec$, which is generated by the model according to a Poisson distribution, and the data label $\labely$, which is given by a Kronecker delta, that is, without label noise.
Here, we assume non-negative observed data and use the Poisson distribution as an elementary distribution for such data (compare restricted Boltzmann machines or sigmoid-belief-networks).
While a Poisson distribution is a natural choice for non-negative data, it also turns out to be mathematically convenient for the derivation of our inference and learning rules.

Note for the model in \cref{eq:PriorK,eq:PriorC,eq:Obs}, that while the normalization of the rows of $\Rgen$ is required for normalized categorical distributions, the normalization of the rows of $\Wgen$ represents an additional assumption of our approach.
By constraining the weights to sum to a constant $\normA$, the model expects contrast normalized data.
If the dimensionality $\dimD$ of the observed data is sufficiently large, we can simply normalize the data such that $\sum_\dimd{}\yd=\normA$ in order to fulfill this constraint with high accuracy.
Denoting the unnormalized data points by $\vec{\tilde{\y}}$, we here assume the normalized data points $\yVec$ to be obtained as follows:
\begin{align}
\yd &= (\normA-\dimD)\frac{\tilde{\y}_\dimd}{\sum_{\dimd^\prime}\tilde{\y}_{\dimd^\prime}}+1. \label{eq:Inputy}
\end{align}

To generate an observation $\yVec$ from the model, we first draw a super class $\classk$ from a uniform categorical distribution $p(\classk)$.
Next we draw a subclass $\subc$ according to the conditional categorical distribution $p(\subc|\classk,\Rgen)$.
Given the subclass, we then sample $\yVec$ from a Poisson distribution and assign to it the label~$\labely$ corresponding to class~$\classk$.
\Cref{eq:PriorK,eq:PriorC,eq:Obs} define a deep mixture model.

\subsection{Maximum Likelihood Learning}
To infer the model parameters $\Theta=(\Wgen,\Rgen)$ of the deep Poisson mixture model \cref{eq:PriorK,eq:PriorC,eq:Obs} for a given set of $\inpN$ independent observed data points \mbox{$\{\yVecN\,\}_{\inpn=1,\ldots,N}$} with $\yVecN\in\!\RRR_{\ge 0}^\dimD$, $\sum_\dimd \ydN = \normA$, and labels $\labelyN$, we seek to maximize the data (log-)likelihood
\begin{equation}
 \mathcal{L}(\Theta) = \log\prod_{\inpn=1}^\inpN p(\yVecN,\labelyN\,|\Theta) = \!\sum_{\inpn=1}^\inpN \log\! \Bigg(\! \sum_{\subc=1}^\subC \!\bigg( \!\! \Big( \prod_{\dimd=1}^\dimD \frac{\Wgencd^{\ydN}e^{-\Wgencd}}{\Gamma (\ydN \! + \! 1)}\Big) \!\!\!\!\sum_{\hphantom{o}\classk \in \labelyN}\!\!\!\frac{\Rgenkc}{\classK} \bigg)\!\! \Bigg).
\label{loglikelihood}
\end{equation}
Here, we assume that some or all of the data come with a label.
For unlabeled data, the summation over $\classk$ is a summation over all possible labels of the given data, that is, $\classk = 1\dots\classK$.
Whereas whenever the label $\labelyN$ is known for a data point $\yVecN$, this sum is reduced to $\classk = \labelyN$, such that only weights $\R[\labelyN\,][\subc]$ contribute for that $\inpn$th data point.

Instead of maximizing the likelihood directly, EM \citep[in the form studied by][]{NealHinton1998} maximizes a lower bound---the free energy---given by:
\begin{equation}
\FF(\ThetaOld,\Theta) =
    \sum\limits_{\inpn=1}^{\inpN} \left\langle \log p(\yVecN,\labelyN,\subc,\classk|\Theta)
    \right\rangle_{\!\inpn} + \mathcal{H}[\ThetaOld],
\label{eq:FreeEnergy}
\end{equation}
where $\langle \, \, \rangle_\inpn$ denotes the expectation under the posterior
\begin{equation}
\E{f(\subc,\classk)}_\inpn = \sum_{\subc=1}^\subC\sum_{\classk=1}^\classK\,p(\subc,\classk|\yVecN,\labelyN,\ThetaOld)\, f(\subc,\classk)
\end{equation}
and  $\mathcal{H}[\ThetaOld]$ is an entropy term only depending on parameter values held fixed during the optimization of $\FF$ w.r.t.\ $\Theta$.
For our model, the free energy as a lower bound of the log-likelihood reads
\begin{equation} 
  \begin{aligned}
&\mathcal{F}(\ThetaOld,\Theta) = \sum_{\inpn,\subc,\classk} p(\subc,\classk|\yVecN,\labelyN,\ThetaOld) \, \bigg(\sum_{\dimd=1}^\dimD \! \Big( \ydN\log\!\big( \Wgencd\big) - \Wgencd - \log\!\big(\Gamma (\ydN \! + \! 1)\!\big) \!\Big) \\[-6pt]
&\hphantom{\mathcal{F}(\ThetaOld,\Theta) = \sum_{\inpn,\subc,\classk} p(\subc,\classk|\yVecN,\labelyN,\ThetaOld) \cdot \bigg(} \, + \log\!\big(\Rgenkc\big) - \log\!\big(\classK\big) \!\bigg) + \mathcal{H}[\ThetaOld].
  \end{aligned}
\label{eq:GenFreeEnergy}
\end{equation}

The EM algorithm optimizes the free energy by iterating two steps:
First, given the current parameters $\ThetaOld$, the relevant expectation values under the posterior are computed in the E-step.
Given these posterior expectations, $\FF(\ThetaOld,\Theta)$ is then maximized w.r.t.\ $\Theta$ in the M-step.
Iteratively applying E- and M-steps locally maximizes the data likelihood.

\paragraph{M-step.}
The parameter update equations of the model can canonically be derived by maximizing the free energy \cref{eq:GenFreeEnergy} under the given boundary conditions of \cref{eq:PriorC,eq:Obs}.
By using Lagrange multipliers for constrained optimization, we obtain after straightforward derivations:
\begin{align}
	\Wgencd &= \normA \frac{\sum_\inpn p(\subc|\yVecN,\labelyN,\ThetaOld)\ydN}{\sum_{\dimd'}\sum_\inpn p(\subc|\yVecN,\labelyN,\ThetaOld)\yN[\dimd']}, \label{eq:MstepW} \\[4pt]
	\Rgenkc &= \frac{\sum_\inpn p(\classk|\subc,\labelyN,\ThetaOld)p(\subc|\yVecN,\labelyN,\ThetaOld)}{\sum_{\subc'}\sum_\inpn p(\classk|\subc',\labelyN,\ThetaOld)p(\subc'|\yVecN,\labelyN,\ThetaOld)}.
\label{eq:MstepR}
\end{align}
For details please refer to \cref{sec:AppEMDetails}.

\paragraph{E-step.}
For the hierarchical mixture model, the required posteriors over the unobserved latents in \cref{eq:MstepW,eq:MstepR} can be efficiently computed in closed-forms in the E-step. Due to an interplay of the used Poisson distribution and the constraint for $\Wgen$ of \cref{eq:Obs}, the equations greatly simplify, and can be
shown to follow a softmax function with weighted sums over inputs $\ydN$ and $\ukN$ as arguments (see \cref{sec:AppEMDetails}):
\begin{align}
p(\subc|\yVecN,\labelyN,\ThetaOld) &= \frac{\exp(\IcN\,)}{\sum_{c^\prime}\exp(\Ic[\subc'][(\inpn)])}\text{, with} \label{eq:PostC}\\[6pt]
%
\IcN &= \sum_\dimd \log(\Wgencd^{\mathrm{old}})\ydN + \log(\sum_\classk \ukN\, \Rgenkc^{\mathrm{old}})\label{eq:Ic} \\
\ukN &= \left\{ \begin{array}{ll}p(\classk|\labelyN\,)=\delta_{\classk\labelyN} & \textnormal{\footnotesize for labeled data}\\ p(\classk)=\frac{1}{\classK} & \textnormal{\footnotesize for unlabeled data}\end{array}\right.\label{eq:Inputu}
\end{align}
Also note, that the posteriors $p(\subc|\yVec,\labely,\Theta)$ for labeled data and $p(\subc|\yVec,\Theta)$ for unlabeled data only differ in the chosen distribution for $\uk$.

For the E-step posterior over classes $\classk$, we obtain:
\begin{align}
& p(\classk|\subc,\labelyN,\ThetaOld) = \left\{ \begin{array}{ll}p(\classk|\labelyN\,)=\delta_{\classk\labelyN} & \textnormal{\footnotesize for labeled data}\\ p(\classk|\subc,\ThetaOld)=\frac{\Rgenkc^{\mathrm{old}}}{\sum_{\classk'}\Rgen[\classk'][\!\subc]^{\mathrm{old}}} & \textnormal{\footnotesize for unlabeled data}\end{array}\right.\label{eq:PostK}
\end{align}
The expression for unlabeled data makes use of the assumption of a uniform prior in \cref{eq:PriorK}.
Under the assumption of a non-uniform class distribution, the weights $\Rgenkc$ would be weighted by the priors $p(\classk)$, which here simply cancel out.

\paragraph{Probabilistically Optimal Classification.}
Once we have obtained a set of values for model parameters~$\Theta$ by applying the EM algorithm on 
training data, we can use the optimized generative model to infer
the posterior distribution $p(\classk|\yVec,\Theta)$ given a previously 
unseen observation $\yVec$. For our model this posterior is given by
\begin{align}
& p(\classk|\yVec,\Theta) = \sum_\subc \frac{\Rgenkc}{\sum_{\classk'}\Rgen[\classk'][\subc]}p(\subc|\yVec,\Theta).  \label{eq:PostKE}  
\end{align}
While this expression provides a full posterior distribution, the maximum a-posteriori (MAP) value can be used for deterministic classification.

\section[Neural Network]{A Neural Network for Optimal Hierarchical Learning}
\label{sec:NeuralNetwork}
For the purposes of this study, we now turn to the task of finding a neural network formulation that corresponds to learning and inference in the hierarchical generative model of~\cref{SecGenerativeModel}. 
The study of optimal learning and inference with neural networks is a popular research field, and we here follow an approach similar to \citet{LuckeSahani2008}, \citet{NesslerEtAl2009}, \citet{KeckEtAl2012} and \citet{NesslerEtAl2013}.

\subsection{A Neural Network Approximation} 
\label{sec:NNApproximation}

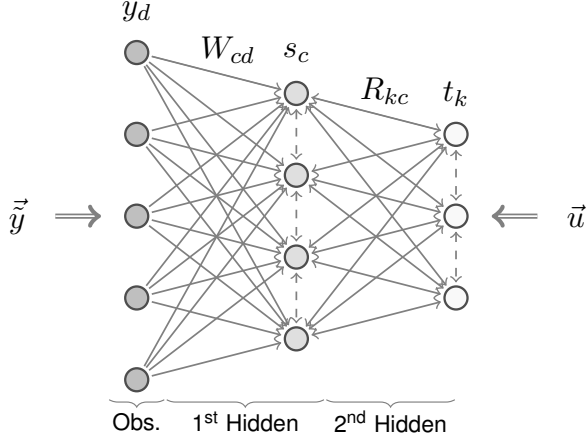
\begin{wrapfigure}[15]{l}{0.50\textwidth}
	\centering
	\resizebox{0.48\textwidth}{!}{
		\begin{adjustbox}{trim=10pt 0pt 10pt 5pt}%
			\begin{tikzpicture}[shorten >=1pt, shorten <= 1pt, draw=black!50, node distance=2.5cm, inner sep=0.333em, outer sep=0.5\pgflinewidth, font=\sffamily]
	\tikzstyle{every pin edge}=[<-,shorten <=1pt]
	\tikzstyle{neuron}=[circle,fill=black!25,color=black!70,thick,minimum size=8pt,inner sep=0pt]
	\tikzstyle{input neuron}=[neuron, draw, fill = gray!50]; 
	\tikzstyle{proc1 neuron}=[neuron, draw, fill = gray!25];
	\tikzstyle{proc2 neuron}=[neuron, draw, fill = gray!5];
	\tikzstyle{annot} = [text width=1em, text centered]
	\def\layerdistance{1.95cm}
	\def\nodedistance{1cm}
	\def\ninput{5}
	\def\nproca{4}
	\def\nprocb{3}
	
	\node[annot] (I0) at (-\layerdistance*0.75,-\nodedistance/2-\ninput/2*\nodedistance) {$\vec{\tilde{y}}$};

	\node[annot] (I) at (0,0.5cm-\nodedistance) {$y_d$};
	\foreach \name / \n in {1,...,\ninput}
			\node[input neuron] (I-\name) at (0,-\n*\nodedistance) {};

	\path (I0) edge [->,double,line width=0.6pt,shorten >=8pt,shorten <=4pt] (I-3);
	
	\node[annot] (P1) at (\layerdistance,0.5cm-\ninput/2*\nodedistance + \nproca/2*\nodedistance - \nodedistance) {$s_c$};
	\foreach \name / \n in {1,...,\nproca}
		\node[proc1 neuron] (P1-\name) at (\layerdistance,-\ninput/2*\nodedistance + \nproca/2*\nodedistance - \n*\nodedistance) {};

	\node[annot] (P2) at (2*\layerdistance,0.5cm-\ninput/2*\nodedistance + \nprocb/2*\nodedistance - \nodedistance) {$t_k$};
	\foreach \name / \n in {1,...,\nprocb}
		\node[proc2 neuron] (P2-\name) at (2*\layerdistance,-\ninput/2*\nodedistance + \nprocb/2*\nodedistance - \n*\nodedistance) {};

	\foreach \source in {1,...,\ninput}
		\foreach \dest in {1,...,\nproca}
			\path (I-\source) edge [->, semithick] (P1-\dest);
	\path (I-1) edge node [right, pos=0.3, yshift=0.2cm] {$W_{cd}$} (P1-1);
					
	\path (P1-1) edge [<->,line width=0.6pt,dashed] (P1-2);
	\path (P1-2) edge [<->,line width=0.6pt,dashed] (P1-3);
	\path (P1-3) edge [<->,line width=0.6pt,dashed] (P1-4);
%

	\path (P2-1) edge [<->,line width=0.6pt,dashed] (P2-2);
	\path (P2-2) edge [<->,line width=0.6pt,dashed] (P2-3);

	\foreach \source in {1,...,\nproca}
		\foreach \dest in {1,...,\nprocb}
			\path (P1-\source) edge [<->, semithick] (P2-\dest);
	\path (P1-1) edge node [right, pos=0.3, yshift=0.2cm] {$R_{kc}$} (P2-1);

	\node[annot] (L) at (2.75*\layerdistance,-\ninput/2*\nodedistance - 1/2*\nodedistance) {$\vec{u}$};

	\path (L) edge [->,double,line width=0.6pt,shorten >=8pt,shorten <= 4pt] (P2-2);
	
	\draw [decorate,decoration={brace,amplitude=3.5pt, mirror},xshift=0pt,yshift=0]
	(-10pt,-\ninput*\nodedistance-6pt) -- (+10pt,-\ninput*\nodedistance-6pt) node [black,midway,sloped, rotate=0, yshift=-8.5pt] 
	{\scriptsize Obs.};
	\draw [decorate,decoration={brace,amplitude=3.5pt, mirror},xshift=0pt,yshift=0]
	(10pt,-\ninput*\nodedistance-6pt) -- (\layerdistance+10pt,-\ninput*\nodedistance-6pt) node [black,midway,sloped, rotate=0, yshift=-8.5pt] 
	{\scriptsize \nth{1} Hidden};
	\draw [decorate,decoration={brace,amplitude=3.5pt, mirror},xshift=0pt,yshift=0]
	(\layerdistance+10pt,-\ninput*\nodedistance-6pt) -- (2*\layerdistance,-\ninput*\nodedistance-6pt) node [black,midway,sloped, rotate=0, yshift=-8.5pt] 
	{\scriptsize \nth{2} Hidden};
\end{tikzpicture}
		\end{adjustbox}
	}
	\caption{Graphical illustration of the hierarchical recurrent neural network.}
	\label{fig:rNN}
\end{wrapfigure}

Consider the neural network in \cref{fig:rNN} with neural activities $\yVec$, $\sVec$ and $\tVec$.
We refer to neurons $\y[][1\ldots\dimD]$\, as the observed layer, the neurons $\Sc[1\ldots\subC]$ make up the first hidden layer, and the neurons $\tk[1\ldots\classK]$ form the second hidden layer.
We assume the values of $\yVec$ to be obtained from a set of unnormalized data points $\vec{\tilde{\y}}$ by \cref{eq:Inputy}, and the label information to be presented as top-down input vector $\uVec{}$ as given in \cref{eq:Inputu}.

Furthermore, we assume the neural activities $\sVec$ and $\tVec$ to be normalized to $B$ and $B'$ respectively (such that $\sum_\dimd\yd=\normA$, $\sum_\classk\uk=1$, $\sum_\subc\Sc=B$, and $\sum_\classk\tk=B^\prime$, with $\normA > \dimD$; $B,B'>0$). 
For the neural weights $(\W,\R)$ of the network---which we distinguish for now from the generative weights ($\Wgen,\Rgen$) of the
mixture model---we consider Hebbian learning with a subtractive synaptic scaling term \citep[see for example][]{AbbottNelson2000}:
\begin{align}
\Delta\Wcd &= \eW (\Sc \yd - \Sc \Wcd) \label{eq:DeltaW}\\[4pt]
\Delta\Rkc &= \eR (\tk \Sc - \tk \Rkc),\label{eq:DeltaR}
\end{align}
where $\eW>0$ and $\eR>0$ are learning rates.
These learning rules are local, can integrate both supervised and unsupervised learning, are highly parallelizable and they result in normalized weights, that we can relate to our generative model as follows:
By taking sums over $\dimd$ and $\subc$ respectively, we observe that the learning dynamics results in $\sum_\dimd\Wcd$ to converge to $\normA$ and $\sum_\subc\Rkc$ to converge to $B$ (due to activities $\yVec$ and $\sVec$ being normalized accordingly).
If we therefore now assume the weights $\W$ and $\R$ to be normalized to $\normA$ and $B$, respectively, we can compute how a given weight adapts with cumulative learning steps.
For small learning rates, we can approximate the weight updates by $\Delta\Wcd = \eW \Sc \yd$ and $\Delta\Rkc = \eR \tk \Sc$ followed by explicit normalization to $\normA$ and $B$, respectively.
Using the superscript $(\inpn)$ to denote the parameter states and activities of the network at the $\inpn$th learning step, we can write the effect of such subsequent weight updates as
\begin{align} 
\Wcd^{(\inpn+1)} = \normA \frac{\WcdN +\, \eW\,\ScN\,\ydN}
  {\sum_{\dimd'}\!\big(\W[\subc][\dimd']^{(\inpn)} +\, \eW\,\ScN\,\y[(\inpn)][\dimd']\,\big)}
  \quad \textnormal{and} \quad
\Rkc^{(n+1)} = B \frac{\RkcN +\, \eR\,\tkN\,\ScN}
  {\sum_{\subc'}\!\big(\R[\classk][\subc']^{(\inpn)} +\, \eR\,\tkN\,\Sc[\subc'][(n)]\,\big) },
  \label[pluralequation]{eq:FP}
\end{align}
where $\ScN =\Sc(\yVecN,\uVecN,\W^{(\inpn)},\R^{(\inpn)})$ denotes the activation of neurons $\Sc$ at the $\inpn$th iteration, which depends on inputs $\yVecN$, $\uVecN$ and the weights $\W^{(\inpn)},\R^{(\inpn)}$.
Similarly, $\tkN=\tk(\sVecN,\uVecN,\R^{(\inpn)})$ depends on $\sVecN$, $\uVecN$, and $\R^{(\inpn)}$.
By iteratively applying \cref{eq:FP} for $\inpN$ times, we can obtain formulas for the weights $\W^{(\inpN)}$ and $\R^{(\inpN)}$---the weights after having learned from $\inpN$ data points.
If learning converges and $\inpN$ is large enough, these can be regarded as the converged weights.
It turns out, that the emerging large nested sums can, at the point of convergence, be compactly rewritten through the use of Taylor expansions and the geometric series.
\Cref{SecOnlineLearning} gives details on the necessary analytical steps.
As a result, we obtain that the following equations must be satisfied for $\W$ and $\R$ at convergence:
\begin{align}
\Wcd \approx \normA \frac{\sum_\inpn \ScN\ydN}{\sum_\dimd \sum_\inpn \ScN\ydN} \qquad \textnormal{and} \qquad \Rkc \approx B \frac{\sum_\inpn \tkN\ScN}{\sum_\subc \sum_\inpn \tkN\ScN}\,.\label[pluralequation]{eq:WRLearning}
\end{align}
\Cref{eq:WRLearning} become exact fixed points for learning in \cref{eq:DeltaW,eq:DeltaR} in the limit of small learning rates $\eW$ and $\eR$ and large numbers of data points $\inpN$.
Given the normalization constraints demanded above, \cref{eq:WRLearning} apply for any neural activation rules for $\Sc$ and $\tk$ as long as learning follows \cref{eq:DeltaW,eq:DeltaR} and as long as learning converges.

For our purpose, we identify $\Sc$ with the posterior probability $p(\subc|\yVec,\labely,\Theta)$ for labeled data and $p(\subc|\yVec,\Theta)$ for unlabeled data given by \cref{eq:PostC,eq:Ic,eq:Inputu} with $\Theta=(\W,\R)$:
\begin{align}
&\ScN := p(\subc|\yVecN,\labelyN,\Theta^{(\inpn)}) = \frac{\exp(\IcN\,)}{\sum_{c^\prime}\exp(\Ic[\subc'][(\inpn)])}\text{, with} \\[6pt]
&\IcN = \sum_\dimd \log(\WcdN)\ydN + \log(\sum_\classk \ukN\,\RkcN)
\end{align}
and $\ukN$ as given by \cref{eq:Inputu}, which incorporates the label information.

Furthermore, we identify $\tk$ with the posterior distribution over classes $\classk$, which for labeled data is $p(\classk|\labely)$ given in \cref{eq:PostK} and for unlabeled data $p(\classk|\yVec,\Theta)$ as given by \cref{eq:PostKE}:
\begin{equation}
\tkN := \left\{\begin{array}{ll}
  p(\classk|\labelyN\,) = \delta_{\labelyN\classk} & ${\footnotesize for labeled data}$\\
  p(\classk|\yVecN,\Theta^{(\inpn)}) = \sum_\subc \! \frac{\RkcN}{\sum_{\classk'} \R[\classk'][\!\subc]^{(\inpn)}} \ScN & ${\footnotesize for unlabeled data}$
\end{array}\right.
\label{eq:NNPosterior}
\end{equation}

The complete set of activation and learning rules, after identifying neural activities $\Sc$ and $\tk$ with the respective posterior distributions, are summarized in \cref{tab:learningrules}:
\begin{table}[!hbt]
	\centering
	{
\small\sffamily
\renewcommand{\arraystretch}{1.4}
\newcommand{\midspace}{\hphantom{abc}}
\setlength{\tabcolsep}{6pt}
\begin{tabular}{@{}l l r}
\toprule
\multicolumn{3}{c}{\bf\sffamily Neural Simpletron}\\
\bottomrule
\multicolumn{2}{l}{\bf\sffamily\scriptsize Input}\\[0pt]
{\scriptsize \hspace{0pt} Bottom-Up: }& $\tilde{\y}_\dimd$ \quad {\scriptsize unnormalized data} & {\scriptsize(T1.1)}\\[2pt]
{\scriptsize \hspace{0pt} Top-Down: }& $\uk = \left\{ \begin{array}{ll}\delta_{\classk\labely} & ${\scriptsize for labeled data}$\\ \frac{1}{\classK} & ${\scriptsize for unlabeled data}$\end{array}\right.$& {\scriptsize(T1.2)}\\
\midrule
\multicolumn{2}{l}{\bf\sffamily\scriptsize Activation Across Layers}\\[0pt]
{\scriptsize \hspace{0pt} Obs. Layer: }& $\yd = (\normA - \dimD) \frac{\tilde{\y}_\dimd}{\sum_{\dimd'}\tilde{\y}_{\dimd'}}+1$& {\scriptsize(T1.3)}\\[3pt]
{\scriptsize\makebox[14pt]{\nth{1}} Hidden:} & \,$\Sc = \frac{\exp(\Ic)}{\sum_{\subc'}\exp(\Ic[\subc'])}${\scriptsize, with}& {\scriptsize(T1.4)}\\[1pt]
 & {\fontsize{9}{0}\;$\Ic=$} {\fontsize{8}{0}$ \sum_\dimd \log(\Wcd)\yd + \log(\sum_\classk \uk \Rkc)$}& {\scriptsize(T1.5)}\\[1pt]
\vspace{3pt}
{\scriptsize \makebox[14pt]{\nth{2}} Hidden:} & \,$\tk = 
\left\{
\begin{array}{ll}
  \uk & ${\scriptsize for labeled data}$\\
  \sum_\subc \! \frac{\Rkc}{\sum_{\classk'} \R[\classk'][\!\subc]} \Sc & ${\scriptsize for unlabeled data}$
\end{array}
\right.$ & {\scriptsize(T1.6)}\\ 
%
\midrule
\multicolumn{2}{l}{\bf\sffamily\scriptsize Learning of Neural Weights} \\[0pt]
{\scriptsize\makebox[14pt]{\nth{1}} Hidden: }& $\Delta \Wcd = \eW(\Sc \yd - \Sc \Wcd)$ & {\scriptsize(T1.7)}\\[3pt]
{\scriptsize\makebox[14pt]{\nth{2}} Hidden: }& $\Delta \Rkc \,= \eR(\tk \Sc - \tk \Rkc)$ {\scriptsize for labeled data} & {\scriptsize(T1.8)}\\[3pt]
\bottomrule

\end{tabular}
}

	\caption{Neural network formulation of probabilistic inference and maximum likelihood learning.}
	\label{tab:learningrules}
\end{table}

By comparing \cref{eq:WRLearning} with the M-step \cref{eq:MstepW,eq:MstepR}, we can now observe, that such neural learning converges to the same fixed points as EM for the hierarchical Poisson mixture model (note, that we set $B=B'=1$ as $\Sc$ and $\tk$ sum to one).
While the identification of $\Wcd$ with $\Wgencd$ at convergence is straightforward, we have to restrict learning of $\Rgenkc$ to labeled data to gain a neural equivalent in $\Rkc$.
In that case $p(\classk|\subc,\labelyN,\ThetaOld) = p(\classk|\labelyN\,)$, which corresponds to our chosen activities $\tk$ for labeled inputs.
(In \cref{sec:SelfLabeling}, we will show a way to loosen up on this restriction by using self-labeling on unlabeled data with high inference certainty.)

In other words, by executing the online neural network of \cref{tab:learningrules}, we optimize the likelihood of the generative model \cref{eq:PriorK,eq:PriorC,eq:Obs}.
The neural activities therein provide the posterior probabilities, which we can, for example, use for classification.
The computation of posteriors is in general a difficult and computationally intensive endeavor, and their interpretation as neural activation rules is usually difficult.
In our case, because of a specific interplay between introduced constraints, categorical distribution and Poisson noise, the posteriors and their neural interpretability greatly simplify, however.

All equations in \cref{tab:learningrules} can directly be interpreted as neural activation or learning rules.
Let us consider an unnormalized data point $\vec{\tilde{\y}}=(\tilde{\y}_1\,,\ldots,\tilde{\y}_\dimD)^T$ as bottom-up input to the network.
Labels are neurally coded as top-down information $\uVec=(\uu[1]\,,\ldots,\uu[\classK])^T$, where only the entry $\uu[\labely]\,$ equals one if $\labely$ is the label, and all other units are zero\footnote{This is sometimes referred to as `one-hot' coding.}.
In the case of unlabeled data, all labels are assumed as equally likely at $1/\classK$.
As first processing step a divisive normalization Eq.~(T1.3) is executed to obtain activations $\yd$. 
Considering Eqs.~(T1.4) and (T1.5), we can interpret $\Ic$ as input to neural unit $\Sc$.
The input consists of a bottom-up and a top-down activation.
The bottom-up input is the standard weighted summation of neural networks $\sum_\dimd \log(\Wcd)\yd$ (note, that we could redefine the weights by $\widetilde{\W}_{\subc\dimd}:=\log\Wcd$). 
Likewise, the top-down input is a standard weighted sum, $\sum_\classk \uk\Rkc$, but affects the input through a logarithm.
Both sums can be computed locally at the neural unit~$\subc$.
The inputs to the hidden units $\Sc$ are then combined using a softmax function, which is also standard for neural networks.
However, in contrast to discriminative networks, the weighted sums and the softmax function are here a direct result from the correspondence to a generative mixture model \citep[compare also][]{JordanJacobs1994}.
The activation of the top layer, Eq.~(T1.6), is either directly given by the top-down input $\uk$, if the data label is know.
Or, for unlabeled data, the inference takes again the form of a weighted sum over bottom-up inputs, which are now the activations $\Sc$ from the middle layer.
Regarding learning, both Eqs.~(T1.7) and (T1.8) are local Hebbian learning equations with synaptic scaling.
The weights of the first hidden layer are updated on all data points during learning, while those of the second hidden layer only learn from labeled input data.

As control of our analytical derivation of \cref{tab:learningrules}, we verified numerically that the local optima of the neural network are indeed also local optima of the EM algorithm.
Note in this respect, that, although neural learning has the same convergence points as EM learning for the mixture model, in finite distances from the convergence points, neural learning follows different gradients, such that the trajectories of the network in parameter space are different from EM.
By adjusting the learning rates in Eqs.~(T1.7) and (T1.8), the gradient directions can be changed in a systematic way without changing the convergence points, which we observed to be beneficial to avoid convergence to shallow local optima.

The equations defining the neural network are elementary, very compact, and contain a total number of only four free parameters: the number of hidden units $\subC$, an input normalization constant $\normA$, and learning rates $\eW$ and $\eR$.
Because of its compactness we call the network {\em Neural Simpletron} (NeSi).

In the experiments in \cref{sec:NumericalExperiments}, we differentiate between four neural network approximations on the basis of \cref{tab:learningrules}. These result from two different approximations of the activations in the first hidden layer, and two different approximations for the activations in the second hidden layer, which gives a total of two by two different networks to investigate.
These approximations in the first and second hidden layer are discussed in the following two subsections respectively.

\subsection{Recurrent, Feedforward and Greedy Learning}

The complete formulas for the first hidden layer, given in Eqs.~(T1.4) and (T1.5), define a recurrent network, that is, a network that combines both bottom-up and top-down information:
The first summation in $\Ic$ incorporates the bottom-up information.
Due to the chosen normalization in Eq.~(T1.3) with a background value of $+1$, all summands in this term are non-negative.
Values of the sum over these bottom-up connections will be high for input data $\yVec$ that was generated by the hidden unit~$\subc$.
The second summation in $\Ic$ incorporates top-down information.
The weighted sum inside the logarithm, which can take the label information into account, will always yield values between zero and one.
Thus, because of the logarithm, this second term is always non-positive and suppresses the activation of the unit.
This suppression is stronger, the less likely it is, that the given hidden unit~$\subc$ belongs to the class of the provided label~$\labely$ (for labeled data) and the less likely it is, that this unit becomes active at all.
Because of these recurrent connections between the first and second hidden layer, we will refer to our method \cref{tab:learningrules} as r-NeSi (`r' for {\em recurrent}) in the experiments.
Whereby, with `recurrent' we do not mean a temporal memory of sequential inputs, but only the direction in which information flows through the network \citep[following, for example,][]{DayanAbbott2001}.

To investigate the influence of such recurrent information in the network, we also test a pure feedforward version of the first hidden layer.
There, we remove all top-down connections by simply discarding the second term in Eq.~(T1.5).
Such a feedforward formulation of the network is equivalent to treating the distribution $p(\subc|\classk,\R)$ in the first hidden layer as a uniform prior distribution $p(\subc)=1/\subC$. 
We will refer to this feedforward network as `ff-NeSi' in the experiments.
Since ff-NeSi is stripped of all top-down recurrence and the fixed points of the second hidden layer now only depend on the activities of the first hidden layer at convergence, it can also be trained disjointly using a greedy layer-by-layer approach, which is customary for deep networks.

\subsection{Self-Labeling}
\label{sec:SelfLabeling}

So far, we trained the top layer of NeSi completely supervised by updating the weights in Eq.~(T1.8) only on labeled data.
When labeled data is sparse, it could be beneficial to also make use of unlabeled data in this layer.
We can do so, by letting the network itself provide the missing labels \citep[a procedure often termed `self-labeling', see, for example,][]{Lee2013,TrigueroEtAl2015}.
The availability of the full posterior distribution in the network (Eq.~T1.6 for unlabeled data) herein allows us to selectively only use those inferred labels where the network shows a very high classification certainty.
As index for decision certainty we use the `Best versus Second Best'~($\BvSB$) measure on $\tk$, which is simply the absolute difference between the most likely and the second most likely prediction.
Such a measure gives a sensible indicator for high skewness of the distribution towards a single class \citep{JoshiEtAl2009}.
If the $\BvSB$ lies above some threshold parameter~$\vartheta$, which we treat as additional free parameter, we can approximate the full posterior in $\tk$ by the MAP estimate.
In that case, we set $\tk \rightarrow \mathrm{MAP}(\tk)$, such that $\tk$ for unlabeled data now holds the 'one-hot' coded inferred label information, with which we update the top layer in the usual fashion using Eq.~(T1.8).

This specific manner of using inferred labels in the neural network is again not imposed ad hoc, but can be derived from the underlying generative model by considering the M-step \cref{eq:MstepR} for unlabeled data.
When in the generative model the posterior $p(\classk|\yVec,\Theta)=\sum_\subc p(\classk|\subc,\Theta)p(\subc|\yVec,\Theta)$ comes close to a hard max, it must be that $p(\subc|\yVec,\Theta)$ is only for those units~$\subc$ dominantly at high values that belong to the same class.
For these units, we can then replace $p(\classk|\subc,\Theta)$ by the MAP estimate in close approximation.
We can therefore rewrite the products in \cref{eq:MstepR} for unlabeled data as
\begin{equation}
  p(\classk|\subc,\Theta)p(\subc|\yVecN,\Theta) \approx \delta_{\classk\tilde{l}}\,p(\subc|\yVecN,\Theta) \quad \forall \inpn\in\inpN: p(\classk|\yVecN,\Theta) \approx \delta_{\classk\tilde{l}},
\label{eq:HighCertainty}
\end{equation}
with the inferred label~$\tilde{\labely}$.
Here, for all data points $\inpn\in\inpN$ with high classification certainty, $p(\subc|\yVecN,\Theta)$ acts as a filter, such that only those terms contribute, where $p(\classk|\subc,\Theta)$ is close to a hard max.
With this approximation, we can replace the dependency of the first factor in \cref{eq:HighCertainty} on specific units~$\subc$ by a common dependency on all units that are connected to unit~$\classk$ (as the inferred label $\tilde{\labely}$ depends on all those units).
These results we are then able to translate again into neural learning rules, where the top layer activation is only dependent on the combined input to that unit, as done above.

We mark those NeSi networks where we use self-labeling in the top layer with~`\textsuperscript{+}' (that is, `r\textsuperscript{+}-NeSi' and `ff\textsuperscript{+}-NeSi').
Although we here use the MAP estimate of $\tk$ during training, because of the validity of \cref{eq:HighCertainty} at high inference certainty, we are still learning in the context of the generative model \cref{eq:PriorK,eq:PriorC,eq:Obs}.
Thus, we still keep the full posterior distribution in $\tk$ for inference, as well as all identifications of \cref{sec:NNApproximation}.

\section{Numerical Experiments}
\label{sec:NumericalExperiments}

We apply an efficiently scalable implementation%
\footnote{
We use a python 2.7 implementation of the NeSi algorithms, which is optimized using Theano to execute on NVIDIA GeForce GTX TITAN Black and TITAN X GPUs.
Details can be found in \cref{AppParallelization}. 
We have provided the source code and scripts for repeating the experiments discussed here along with the submission.}
of our network to three standard benchmarks for classification: the 20~Newsgroups text data set \citep{Lang1995}, the MNIST data set of handwritten digits \citep{LeCunEtAl1998} and the NIST Special Database 19 of handwritten characters \citep{Grother1995}.
To investigate the semi-supervised task, we randomly divide the training parts of the data sets into labeled and unlabeled partitions, where we make sure that each class holds the same number of labeled training examples, if possible.
We repeat experiments for different proportions of labeled data and measure the classification error on the blind test set.
For all such settings, we report the average test error over a given number of independent training runs with new random labeled and unlabeled data selection.
Details on parallelization and weight initialization can be found in \cref{AppTrainingDetails}. Detailed statistics of the obtained results are given in \cref{AppResults}.

\subsection{Parameter Tuning}
\label{sec:ParameterTuning}
For the NeSi algorithms, we have four free parameters:
the normalization constant~$\normA$ in the bottom layer, the number of hidden units~$\subC$ and the learning rate $\eW$ in the middle layer, and the learning rate~$\eR$ in the top layer.
When using the optional self-labeling, we have a fifth free parameter~$\vartheta$ as $\BvSB$ threshold, also in the top layer.

To optimize the free parameters in the semi-supervised setting with only few labeled data points, it is customary to use a validation set, which comprises additional labeled data to the available amount of labels in the training set of that given setting (for example, using a validation set of \num{1000}~labeled data points to tune parameters in the setting of \num{100}~labels).
As this procedure does not guarantee that the resulting optimal parameter setting could have also been found with the limited amount of labels in the given setting, such achieved results reflect more of the performance limit of the model than the actual performance when given only very restricted amounts of labeled data.
As already in \cite{ForsterEtAl2015}, we therefore not only train our model on such limited labeled data, but also tune all free parameters in this same setting without any additional labeled data.
This way we make sure, that our results are achievable by using no more labels than provided within each training setting.
Furthermore, using only training data for parameter optimization assures a fully blind test set, such that the test error gives a reliable index for generalization.

To construct the training and validation set for parameter tuning, we regard the setting of 10~labeled training data points per class (that is, 100~labeled data points for MNIST and 200 for 20~Newsgroups).
This is the setting with the lowest number of labels, on which models are generally compared on MNIST.
For simplicity's sake, we take half of this labeled data as validation set (class balanced and randomly drawn) and use the other labeled half plus all unlabeled training data as training set for parameter tuning.
With this data split, we optimize the parameters of the r-NeSi network via a coarse manual grid search.
For the search space, we may consider run time vs.\ performance trade-offs where necessary (for example, with an upper bound on the network size or a lower bound on the learning rates).
Keeping the optimized parameter setting of r-NeSi fixed, we only optimize~$\vartheta$ for r\textsuperscript{+}-NeSi.
For comparison, we keep the same parameter settings for the feedforward networks (ff-NeSi and ff\textsuperscript{+}-NeSi) without further optimization.

Once optimized in this semi-supervised setting, we keep the free parameters fixed for all following experiments.
When evaluating the performance of the networks, we perform repeated experiments with different sets of randomly chosen training labels.
This evaluation scheme is of course only possible with more labels available than used by each single network.
However, this procedure is purely to gather meaningful statistics about the mean and variance of the acquired results, as these can vary based on the set of randomly chosen labels.
As the experiments are performed independently of each other and the parameters are not further tuned based on these results on the test set, it is safe to say, that the acquired results are a statistical representation of the performance of our models given no more than the corresponding number of labels in each setting.

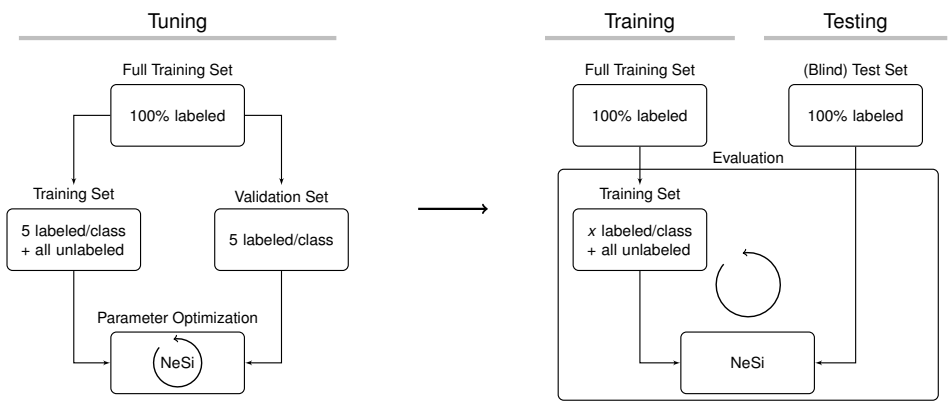
\begin{figure}[!htb]
  \centering
  \resizebox{0.8\columnwidth}{!}{
\tikzstyle{block} = [rectangle, draw, 
    text width=8em, text centered, rounded corners, minimum height=4em]
\tikzstyle{line} = [draw, -latex']
\begin{tikzpicture}[thick,font=\sansmath\sffamily]
  
  \node [] (Tu) at (0,0) {\Large Tuning};
  \draw [line width=3pt, lightgray] ([yshift=-1em,xshift=-8em]Tu.west) -- ([yshift=-1em,xshift=8em]Tu.east);   
  \node [block,below= 3em of Tu, label=Full Training Set] (TuFTS) {$100\%$ labeled};
  \node [block,below left=4em and -2em of TuFTS, label=Training Set] (TuTS) {5 labeled/class \\ + all unlabeled};
  \node [block,below right=4em and -2em of TuFTS, label=Validation Set] (TuVS) {5 labeled/class};
  \node [block,below=12em of TuFTS, label=Parameter Optimization] (TuPO) {NeSi};
  \draw [->,line width=1pt] (TuPO) ++(140:6mm) arc (-220:100:6mm);  
 
  \path [line] (TuFTS.west) -| ([yshift=1.5em]TuTS.north);   
  \path [line] (TuFTS.east) -| ([yshift=1.5em]TuVS.north);   

  \path [line] (TuTS.south) |- (TuPO.west);
  \path [line] (TuVS.south) |- (TuPO.east);


  \node [right of=Tu, node distance=30em] (Tr) {\Large Training};
  \draw [line width=3pt, lightgray] ([yshift=-1em,xshift=-3.5em]Tr.west) -- ([yshift=-1em,xshift=3.5em]Tr.east);
  \node [right of=Tr, node distance=14em] (Te) {\Large Testing};
  \draw [line width=3pt, lightgray] ([yshift=-1em,xshift=-3.5em]Te.west) -- ([yshift=-1em,xshift=3.5em]Te.east);     
  \node [block,below=3em of Tr, label=Full Training Set] (TrFTS) {$100\%$ labeled};
  \node [block,below=3em of Te, label=(Blind) Test Set] (TeTeS) {$100\%$ labeled};
  \node [block,below=4em of TrFTS, label=Training Set] (TrTS) {$x$ labeled/class \\ + all unlabeled};
  \node [block,below right=12em and -1.75em of TrFTS] (TrGE) {NeSi};
  \node [block,below right=8.5em and -8em of Tr, text width=24em, minimum height=15em,label=Evaluation] (TrBox) {};
  \draw [->,line width=1pt] ([yshift=0em,xshift=0em]TrBox) ++(140:8mm) arc (-220:100:8mm);  
  \path [line] (TrFTS) -- ([yshift=1.5em]TrTS.north);
  \path [line] (TrTS.south) |- (TrGE.west);
  \path [line] (TeTeS.south) |- (TrGE.east);

  \draw [->, ultra thick] ([xshift=4.5em, yshift=2em]TuVS.east) -- ([xshift=-5.5em, yshift=2em]TrTS.west);

\end{tikzpicture}  
  }
  \caption{Tuning, training and testing protocol for the NeSi algorithms. During tuning, the free parameters are optimized on a split of the training data into a training and validation set with 5~randomly chosen labeled data points per class in each, and all remaining unlabeled data points in the training set. These data sets with their chosen labels remain fixed during all tuning iterations. With the resulting set of optimized free parameters, the network is then trained on all available training data and labels in the given setting and is evaluated on the fully blind test set. This last training and testing step is repeated with a new, randomly chosen, class balanced set of training labels for multiple independent iterations to gain the mean generalization error of the algorithms.}
  \label{fig:tuningprotocol}
\end{figure}

A more rigorous parameter tuning would also allow for retuning of all parameters for each model and each new label setting, making use of the additional training label information in the stronger labeled settings, which we however refrained to do for our purposes.
The overall tuning, training and testing protocol is shown in \cref{fig:tuningprotocol}.

\subsection{Document Classification (20~Newsgroups)}\label{sec:20Newsgroups}
The 20~Newsgroups data set in the `bydate' version consists of \num{18774} newsgroup documents of which \num{11269} form the training set and the remaining \num{7505} form the test set.
Each data vector comprises the raw occurring frequencies of \num{61188} words in each document.
We preprocess the data using only tf-idf weighting \citep{Sparck1972}. No stemming, removals of stop words or frequency cutoffs were applied.
The documents belong to \num{20} different classes of newsgroup topics that are partitioned into six different subject matters (`comp', `rec', `sci', `forsale', `politics' and `religion').
We show experiments for both classification into subject matter (6 classes) as well as the more difficult full 20-class problem.

\subsubsection{Parameter Tuning on 20~Newsgroups}
In the following, we give a short overview over the parameter tuning on the 20~Newsgroups data set. 
We use the procedure described in \cref{sec:ParameterTuning} to optimize the free parameters of NeSi using only 200~labels in total, while keeping a fully blind test set.
The parameters are optimized with respect to the more common 20-class problem, and we then keep the same parameter setting also for the easier 6-class task.
We allowed training time over 200 iterations over the whole training set and restricted the parameters in the grid search such that convergence was given within this limitation.

\paragraph{Hidden Units.}
Following the above tuning protocol for 20~Newsgroups (20 classes) results in a best performing architecture of $\dimD$--$\subC$--$\classK$\,=\,61188--20--20, that is, the complete setting $\subC$\,=\,$\classK$\,=\,20.
Generally we would expect, that the overcomplete setting $\subC > \classK$ would allow for more expressive representations.
This is indeed the case for the 6-class problem ($\classK=6$) for which we find that $\subC=20$ (61188--20--6) is the best setting but more middle-layer classes were not beneficial for the 20-class problem.
Using more than 20~middle layer units ($\subC > 20$) for $\classK = 20$ problem could be hindered here by the high dimensionality of the data relative to the number of available training data points as well as the prominent noise when taking all words of a given document into account.

\paragraph{Normalization.}
Because of the introduced background value of +1 (see Eq.~T1.3), the normalization constant~$\normA$ has a lower bound in the dimensionality of the input data $\dimD=\num{61188}$.
For very low values $\normA \gtrsim \dimD$, the model is unable to differentiate the observed patterns from background noise.
At the other extreme, at $\normA \rightarrow \infty$, the softmax function will converge to a winner-take-all maximum function.
The optimal value lies in between, closely after the system is able to differentiate all classes from background noise but when the normalization is still low enough to allow for a broad softmax response.
For all our experiments on the 20~Newsgroups data set we chose (following the tuning protocol) $\normA=\num{80000}$ (that is, $\normA/\dimD \approx 1.31$).

\paragraph{Learning Rates.}
A relatively high learning rate in the first hidden layer ($\eW = 5 \times \subC/\inpN$), coupled with a much lower learning rate in the second hidden layer ($\eR = 0.5 \times \classK/L$), yielded the best results on the validation set. 
Especially the high value for $\eW$ seems to have the effect of more efficiently avoiding shallow local optima, which exist, again, due to noise and the high dimensionality of the data compared to the relatively low number of training samples.
The different learning rates for $\eW$ and $\eR$ mean that the neural network follows a gradient markedly different from an EM update.
This suggests, that the neural network allows for improved learning compared to the EM updates it was derived from.

Note, that in practice we use normalized learning rates. The factor $\subC/\inpN$ for the first hidden layer and $\classK/L$ for the second hidden layer represents the average activation per hidden unit over one full iteration over a data set of $\inpN$ data points with $L$ labels.
Tuning not the absolute learning rate but the proportionality to this average activation helps to decouple the optimum of the learning rates from the network size ($\subC$ and $\classK$) and the amount of available training data and labels ($\inpN$ and $L$).

\paragraph{BvSB Threshold.}
Given the optimized values of the other free parameters, we found that introducing the additional self-labeling for unlabeled data is not helpful and even harmful for the 20~Newsgroups data set.
Since even in the settings with only very few labeled data points, the number of provided labels per middle layer hidden unit is already sufficiently large, the usage of inferred labels only introduces destructive noise.
The self-labeling will show to be more useful in scenarios where the number of hidden units surpasses the number of available labeled data points greatly (like for MNIST, \cref{sec:MNIST}, and NIST, \cref{sec:NIST}).

\subsubsection{Results on 20~Newsgroups (6 classes)}
We start with the easier task of subject matter classification, where the twenty newsgroup topics are partitioned into six higher-level groups that combine related topics (`comp', `rec', etc.).
The optimal architecture for 20~Newsgroups (20 classes) on the validation set was given in the complete setting, where $\subC$\,=\,$\classK$\,=\,20.
At first glance, this seems like no subclasses were learned and that the split in the middle layer was primarily guided by class labels.
However, also for classification of subject matters (6 classes), where only labels of the six higher-level topics were given, we observed the setting with $\subC$\,=\,20 units (61188-20-6) to be far superior to the complete setting with architecture 61188-6-6 (see \cref{tab:6News-results}).
This suggests, that the data structure of 20~subclasses determines the optimal architecture of the NeSi network and not the number of label classes (see also \cref{sec:MNIST,sec:NIST}).
In our experiments we furthermore observed the feedforward network, which learns completely unsupervised in the middle layer, to still achieve a similar performance as the recurrent \mbox{r-NeSi} network.
This shows, that the NeSi networks are able to recover individual subclasses of the newsgroups data independently of the label information.
\begin{table}[!h]
  \centering
  {
\fontsize{8}{10} \sffamily \sansmath
\renewcommand{\arraystretch}{1.3}
\newcommand{\midspace}{\hphantom{abc}}
\setlength{\tabcolsep}{14pt}
\scriptsize
\begin{tabular}{@{}l r@{}c@{}l r@{}c@{}l r@{}c@{}l r@{}c@{}l}
\toprule
 & \multicolumn{6}{@{}c}{\hspace{1.0em}ff-NeSi} & \multicolumn{6}{@{}c}{\hspace{1.0em}r-NeSi} \\
\#labels & \multicolumn{3}{@{}c}{\hspace{1.3em}$\subC=6,\,\classK=6$} & \multicolumn{3}{@{}c}{\hspace{1.3em}$\subC=20,\,\classK=6$}  & \multicolumn{3}{@{}c}{\hspace{1.3em}$\subC=6,\,\classK=6$} & \multicolumn{3}{@{}c}{\hspace{1.3em}$\subC=20,\,\classK=6$}\\
\midrule
200 & 41.66 & $\,\pm\,$ & 1.21 & 14.23 & $\,\pm\,$ & 0.45 & 39.02 & $\,\pm\,$ & 1.49 & 14.21 & $\,\pm\,$ & 0.42 \\
800 & 40.41 & $\,\pm\,$ & 1.31 & 14.04 & $\,\pm\,$ & 0.48 & 39.54 & $\,\pm\,$ & 1.64 & 14.58 & $\,\pm\,$ & 0.75 \\
2000 & 42.31 & $\,\pm\,$ & 0.72 & 14.26 & $\,\pm\,$ & 0.47 & 40.05 & $\,\pm\,$ & 0.64 & 13.44 & $\,\pm\,$ & 0.43 \\
11269 & 41.85 & $\,\pm\,$ & 0.90 & 14.95 & $\,\pm\,$ & 0.73 & 36.56 & $\,\pm\,$ & 2.09 & 13.26 & $\,\pm\,$ & 0.35 \\
\bottomrule
\end{tabular}
\setlength{\tabcolsep}{6pt}
}

  \caption{Test error for 10~independent runs on the 20~Newsgroups data set, when classes are combined by their corresponding subject matters (classification into $\classK=6$ classes). Here the overcomplete setting ($\subC > \classK$) shows best results, where the network is able to learn the 20 individual subclasses present in the data.}
  \label{tab:6News-results}
\end{table}

\subsubsection{Results on 20~Newsgroups (20 classes)}
\label{sec:Results20News}
\begin{figure}[!tbh]
  \includegraphics[width=\textwidth]{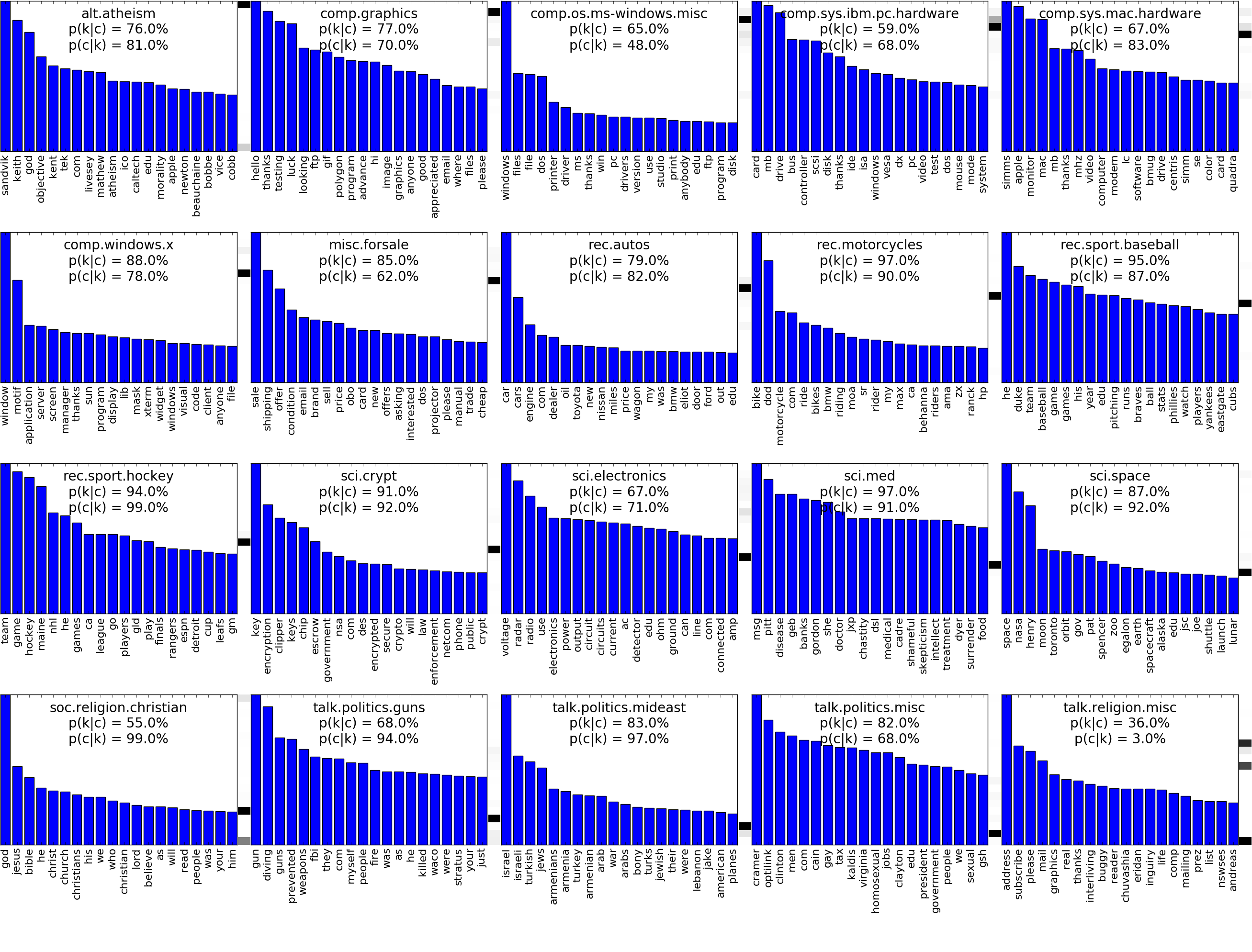}
  \vspace{-18pt}
  \caption{Example of learned weights by the r-NeSi algorithm in the semi-supervised setting of 800~labels.
  Shown are the 20~features with the highest learned tf-idf occurrence frequency for each of the 20~hidden units as bar plot (scaled relatively to the most likely feature).
  Columns next to each field show the corresponding learned class assignment.
  Each field is labeled by the class~$\classk$ with the highest probability $p(\classk|\subc)$ for that field $\subc$.
  For that most likely class, the probabilities $p(\classk|\subc)$ and $p(\subc|\classk)$ are given.
  }
  \label{fig:20News-weights}
\end{figure}

We now continue with the more challenging 20-class problem ($\classK=20$).
Here, we investigate semi-supervised settings of \num{20}, \num{40}, \num{200}, \num{800} and \num{2000} labels in total---that is \num{1}, \num{2}, \num{10}, \num{40} and \num{100} labels per class---as well as the fully labeled setting.
For each setting, we present the mean test error averaged over \num{100} independent runs and the standard error of the mean (SEM).
On each new run, a new set of class balanced labels is chosen randomly from the training set.
We train our model on the full 20-class problem without any feature selection.
An example of some learned weights of r-NeSi is shown in \cref{fig:20News-weights}.

To the best of our knowledge, most methods that report performance on the same benchmark do consider easier tasks: They either break the task into binary classification between individual or merged topics \citep[such as][]{ChengEtAl2006, KimEtAl2014, WangEtAl2012, ZhuEtAl2003}, and/or perform feature selection \citep[for instance,][]{SrivastavaEtAl2013, Settles2011} for classification.
There are however works that are compatible with our experimental setup \citep{LarochelleBengio2008,RanzatoSzummer2008}.
A hybrid of generative and discriminative RBMs (HDRBM) trained by \citet{LarochelleBengio2008} uses stochastic gradient descent to perform semi-supervised learning.
They report results on 20~Newsgroups for both supervised and semi-supervised setups.
In the fully labeled setting, all their model- and hyperparameters are optimized using a validation set of \num{1691} examples with the remaining \num{9578} in the training set.
In the semi-supervised setup \num{200} examples were used as validation set with \num{800} labeled examples in the training set.
To reduce the dimensionality of the input data, they only used the \num{5000} most frequent words.
The classification accuracy of the method is compared in \cref{tab:20News-results}.
\begin{table}[!h]
  \centering
  {
\fontsize{8}{10} \sffamily \sansmath
\renewcommand{\arraystretch}{1.3}
\newcommand{\midspace}{\hphantom{abc}}
\setlength{\tabcolsep}{14pt}
\scriptsize
\begin{tabular}{@{}l r@{}c@{}l r@{}c@{}l c@{}}
\toprule
\#labels & \multicolumn{3}{@{}c}{ff-NeSi} & \multicolumn{3}{@{}c}{r-NeSi} & HDRBM	\\
\midrule
20  & 70.64 & $\,\pm\,$ & 0.68 $^{(*)}$ & \bfseries 68.68 & $\,\pm\,$ & 0.77 $^{(*)}$\\
40  & 55.67 & $\,\pm\,$ & 0.54 $^{(*)}$ & \bfseries 54.24 & $\,\pm\,$ & 0.66 $^{(*)}$\\
200  & 30.59 & $\,\pm\,$ & 0.22 & \bfseries 29.28 & $\,\pm\,$ & 0.21 \\
800  & 28.26 & $\,\pm\,$ & 0.10 &  \bfseries 27.20 & $\,\pm\,$ & 0.07 & 31.8	$^{(*)}$\\
2000 & 27.87 & $\,\pm\,$ & 0.07 &  \bfseries 27.15 & $\,\pm\,$ & 0.07 \\
11269 & 28.08 & $\,\pm\,$ & 0.08 &  27.28 & $\,\pm\,$ & 0.07 & \bfseries 23.8	\phantom{$^{(*)}$} \\
\bottomrule
\end{tabular}
\setlength{\tabcolsep}{6pt}
}

  \caption{Test error on 20~Newsgroups for different label-settings using the feedforward and the recurrent Neural Simpletrons.
  We differentiate here between settings with different amounts of labels available during training.  
  For results marked with $^{(*)}$, the free parameters of the model were optimized using additional labels: NeSi used the same parameter setting in all experiments on 20~Newsgroups, which was tuned with 200~labels in total; HDRBM used 1000~labels in total for tuning in the semi-supervised setting.}
  \label{tab:20News-results}
\end{table}

Here, the recurrent and feedforward networks produce very similar results, with a small advantage to the recurrent networks.
This small advantage could however also be explained by a bias in our tuning procedure, where the parameters are specifically optimized for the recurrent model.
In comparison with HDRBM, ff-NeSi and r-NeSi both achieve better results than the competing model for the semi-supervised setting.
Both algorithms are still better with down to $200$~labels, even though HDRBM uses more labels for training and additional labels for parameter tuning.
Performance only very significantly decreases when going down even further to only one or two labels per class for training (note, that the parameters were actually tuned using 200~labels in total).

\subsubsection{Optimization in the Fully Labeled Setting}
\label{sec:20NewsFullyLabeled}
In the fully labeled setting, the HDRBM outperforms the shown NeSi approaches significantly.
However, we have so far used one parameter tuning fixed for all settings.
We can further optimize for a specific setting, here the fully labeled one.
In that setting, we can still gain a larger benefit out of the recurrence of r-NeSi:
Changing its initialization procedure from $\Rkc = 1/\subC$ to $\Rkc = \delta_{\classk\subc}$ helped to avoid shallow local optima and reached a test error of $(17.85 \, \pm \, 0.01)\%$.
This initialization fixes the class $\classk$ of subclass $\subc$ to a single specific class by setting all connections between the first and second hidden layer to other classes to zero.
Training with such a weight initialization is however only useful when very large amounts of labeled data are available.
The top-down label information is then an important mechanism to make sure, that the middle layer units learn the appropriate representation of their respective fixed class (for example, that a middle layer unit that is fixed to class `alt.atheism' mainly, or exclusively, learns from data belonging to that class).
So, instead of first learning representations in the middle layer purely from the data and then learning the classes with respect to these representations from the labels, like the (greedy) ff-NeSi, the r-NeSi algorithm is able to also conversely shape their middle layer representations in relation to their probability to belong to the class of the presented data point.

To decide between this initialization procedure in the fully labeled setting and our standard one, we here used the fully labeled training set during parameter tuning (again with a half/half split into training and validation set).
With the better avoidance of shallow optima by this initialization, lower learning rates $\eW$ were now more beneficial ($\eR$ drops out as free parameter, as the top layer remains fixed).
A coarse manual grid search in this setting resulted in optimal parameter values at $\normA = \num{90000}$ ($\normA/\dimD \approx 1.47$) and $\eW = 0.02$ (which we chose as lowest search value to restrict computational time), while keeping $\subC=20$.
These results also show, that parameter optimization based on each individual label setting (instead of just on the weakliest labeled setting) and changing the initialization procedure based on label-availability could potentially lead to better parameter settings and stronger performance also in the other settings.

\subsection{Handwritten Digit Recognition (MNIST)}
\label{sec:MNIST}
The MNIST data set consists of \num{60000} training and \num{10000} testing data points of \num{28}$\times$\num{28} images of gray-scale handwritten digits which are centered by pixel mass.
We perform experiments in the semi-supervised setting using \num{10}, \num{100}, \num{600}, \num{1000} and \num{3000} labels in total, which are randomly and class balanced chosen from the 10~classes.
Additionally, we consider the setting of a fully labeled training set.

\subsubsection{Parameter Tuning on MNIST}
We here give a short overview over the parameter tuning on the MNIST data set. 
We again use the tuning procedure described in \cref{sec:ParameterTuning} to optimize all free parameters of NeSi using only 100~labels in total from the training data, keeping a fully blind test set.
We allowed training time over 500 iterations over the whole training set and restricted the parameters in the grid search such that convergence was given within this limitation.

\paragraph{Hidden Units.}
Contrary to the 20~Newsgroups data set, for MNIST the validation error generally decreased with an increasing number of hidden units.
We therefore used $\subC=\num{10000}$ for all our experiments for both the feedforward and the recurrent networks, which we set as upper limit for network size as a good trade-off between performance and required compute time.
However, with such many hidden units on a training set of \num{60000} data points, and with as few as only \num{10} labeled training samples in total, overfitting effects have to be taken into consideration.
We discuss these more deeply in \cref{sec:EarlyStopping,sec:OverfittingControl}.
In general, we encountered an increase in error rates on prolonged training times only for the r-NeSi algorithm in the semi-supervised settings when no self-labeling was used.
For this case only, we devised and used a stopping criterion based on the likelihood of the training data.

\paragraph{Normalization.}
The dependence of the validation error on the normalization constant~$\normA$ shows similar behavior as for the 20~Newsgroups data set. 
Following a screening according to the tuning protocol, the setting of $\normA=900$ (that is, $\normA/\dimD \approx 1.15$) was chosen.

\paragraph{Learning Rates.}
While a high learning rate can be used to overcome shallow local optima, a lower learning rate will in general yield better results with the downside of a longer training time until convergence.
As trade-off between performance and training time, we chose $\eW = 0.2 \times \subC/\inpN$ and $\eR = 0.2 \times \classK/L$ for all experiments on MNIST.
Since for networks using self-labeling the number of effectively used labels $L$ approaches $\inpN$ over time, we scale the learning rate $\eR$ for those system with $\classK/\inpN$ instead of $\classK/L$, that is $\eR = 0.2 \times \classK/\inpN$ for r$^+$- and ff$^+$-NeSi.

\paragraph{BvSB Threshold.}
With $\subC = \num{10000}$ and only \num{50}~labels in total in the training set during parameter tuning, there is only a single label per \num{200} middle layer fields available to learn their respective classes.
In this setting, using self-labeling on unlabeled data as described in \cref{sec:SelfLabeling}, decreased the validation error significantly over the whole tested regime of $\vartheta \in [0.1,0.2,\dots,0.9]$.
We chose $\vartheta = 0.6$ as the optimal value.

\subsubsection{A Likelihood Criterion For Early Stopping}
\label{sec:EarlyStopping}
Training of the first layer in the feedforward network is not influenced by the state of the second layer, and is therefore independent of the number of provided labels.
This is no longer the case for the recurrent network.
A low number of labels can lead to overfitting effects in r-NeSi when the number of hidden units in the first hidden layer is substantially larger than the number of labeled data points.
However, when using the inferred labels for training in the r\textsuperscript{+}-NeSi network such overfitting effects will vanish again.

Since learning in our network corresponds to maximum likelihood learning in a hierarchical generative model, a natural measure to define a criterion for early stopping can be based on monitoring of the log-likelihood, which is given by \cref{loglikelihood} (replacing the generative weights $(\Wgen,\Rgen)$ by the weights $(\W,\R)$ of the network).
As soon as the scarce labeled data starts overfitting the first layer units as a result of top-down influence in $\Ic$ (compare Eq.~T1.5), the log-likelihood computed over the whole training data is observed to decrease.
This declining event in data likelihood can be used as stopping criterion to avoid overfitting without requiring additional labels.

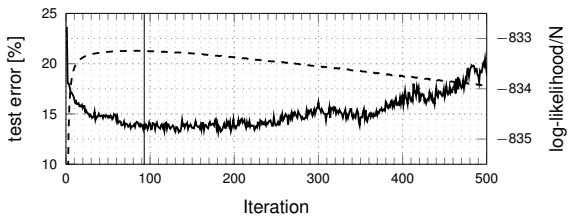
\begin{wrapfigure}{l}{0.5\textwidth}
	\resizebox{0.5\textwidth}{!}{
\pgfplotsset{
	grid style={dotted,gray},
	minor grid style={dotted,lightgray},
  tick label style = {font=\tiny\sansmath\sffamily},
  legend style = {font=\sansmath\sffamily},
  xlabel style = {font=\sansmath\sffamily},
  ylabel style = {font=\sansmath\sffamily},
}	
\begin{tikzpicture}
	\tikzset{mark size={1.0}}
	\newcommand\width{\linewidth} 
	\newcommand\height{0.17 * \textheight}	
	\begin{axis}[
		colormap access=direct,
		width = \width,
		height = \height,
		xmin=0,
		xmax=500,
		minor xtick = {0,10,...,500},
		scaled x ticks = false,
		xlabel={\scriptsize\sffamily Iteration},
		xlabel near ticks, xticklabel pos=lower,
		ymin = 10,
		ymax = 25,
		ytick = {0,5,...,100},
		minor ytick = {5,10,...,95},
		ylabel={\scriptsize\sffamily test error [\%]},
		ylabel near ticks, yticklabel pos=left,
    axis y line*=left,
		grid = both,
		]
		\addplot+ [thick, mark=None, color=black] table[x=X, y=Y1] {./figs/MNIST-likelihood.txt};
    \draw (axis cs:93,\pgfkeysvalueof{/pgfplots/ymin}) -- (axis cs:93,\pgfkeysvalueof{/pgfplots/ymax});
  \end{axis}
	\begin{axis}[
		colormap access=direct,
		width = \width,
		height = \height,
		xmin=0,
		xmax=500,
		hide x axis,
		ymin = -835.5,
		ymax = -832.5,
		ylabel={\scriptsize\sffamily log-likelihood/N},
		ylabel near ticks, yticklabel pos=right,
    axis y line*=right,
		grid = both,
		]
		\addplot+ [thick, dashed, mark=None, color=black] table[x=X, y=Y2] {./figs/MNIST-likelihood.txt};
  \end{axis}  
\end{tikzpicture}
	}
	\vspace{-16pt}
  \caption{Evolution of test error (solid) and log-likelihood (dashed) in r-NeSi during training. Both show a strong negative correlation. The vertical line denotes the stopping point.}
	\label{fig:MNIST-likelihood}
\end{wrapfigure}
\Cref{fig:MNIST-likelihood} shows an example of the evolution of the average log-likelihood per data point during training compared to the test error. 
For experiments over a variety of network sizes, we found strong negative correlations of $\left<\mathrm{PPMCC}\right> = -0.85 \pm 0.1$.
To smooth-out random fluctuations in the likelihood, we compute the centered moving average over \num{20}~iterations and stop as soon as this value drops below its maximum value by more than the centered moving standard deviation.
The test error in \cref{fig:MNIST-likelihood} is only computed for illustration purposes.
In our experiments we solely used the moving average of the likelihood to detect the drop event and stop learning. 
In our control experiments on MNIST, we found that the best test error generally occurred some iterations after the peak in the likelihood (compare \cref{fig:MNIST-likelihood}), which we however for simplicity not exploited for our reported results.

\subsubsection{Results on MNIST}
\label{sec:ResultsMNIST}
\Cref{tab:MNIST-results} shows the results of the NeSi algorithms on the MNIST benchmark. As the NeSi model has no prior knowledge about spatial relations in the data, the given results are invariant to pixel permutation.
As can be observed, the recurrent networks (r-NeSi) result in significantly lower classification errors than the feedforward networks (ff-NeSi) in the fully and the weakliest labeled settings.
In between those extrema, we find a regime where the feedforward networks do not only catch up to the recurrent networks but even perform slightly better.
In this highly over-complete setting, we now also see a significant gain in performance for the semi-supervised settings with the additional self-labeling (ff\textsuperscript{+}-NeSi and r\textsuperscript{+}-NeSi).
With these additional inferred labels, the feedforward network surpasses the recurrent version also in the settings with very few labels, down to a single label per class.
For this last setting however, we had to increase the training time to 2000~iterations to assure convergence, since learning in the top layer with a single label per class per iteration is very slow when not adjusting the learning rate.

\begin{table}[!t]
  \centering
  {
\fontsize{8}{10} \sffamily \sansmath
\renewcommand{\arraystretch}{1.3}
\newcommand{\midspace}{\hphantom{abc}}
\setlength{\tabcolsep}{12pt}
\scriptsize
\begin{tabular}{@{}l r@{}c@{}l r@{}c@{}l r@{}c@{}l r@{}c@{}l}
\toprule
\#labels & \multicolumn{3}{@{}c}{ff-NeSi} & \multicolumn{3}{@{}c}{r-NeSi} & \multicolumn{3}{@{}c}{ff\textsuperscript{+}-NeSi} & \multicolumn{3}{@{}c}{r\textsuperscript{+}-NeSi} \\
\midrule
10   & 55.46 &$\,\pm\,$& 0.57  $^{(*)}$ & 29.61 &$\,\pm\,$& 0.57  $^{(*)}$ & \bfseries 10.91 &$\,\pm\,$& 0.86  $^{(*)}$ & 17.90 &$\,\pm\,$& 0.89  $^{(*)}$\\
100  & 19.08 &$\,\pm\,$& 0.26 & 12.43 &$\,\pm\,$& 0.15 & \bfseries 4.96 &$\,\pm\,$& 0.08 &\bfseries 4.93 &$\,\pm\,$& 0.05\\
600  &  7.27 &$\,\pm\,$& 0.05 &  6.94 &$\,\pm\,$& 0.05 &\bfseries 4.08 &$\,\pm\,$& 0.02 &  4.34 &$\,\pm\,$& 0.01\\
1000 &  5.88 &$\,\pm\,$& 0.03 &  6.07 &$\,\pm\,$& 0.03 &\bfseries 4.00 &$\,\pm\,$& 0.01 &  4.26 &$\,\pm\,$& 0.01\\
3000 &  4.39 &$\,\pm\,$& 0.02 &  4.68 &$\,\pm\,$& 0.02 &\bfseries 3.85 &$\,\pm\,$& 0.01 &  4.05 &$\,\pm\,$& 0.01\\
60000&  3.27 &$\,\pm\,$& 0.01 &\bfseries 2.94 &$\,\pm\,$& 0.01 &  3.27 &$\,\pm\,$& 0.01 &\bfseries 2.94 &$\,\pm\,$& 0.01\\
\bottomrule
\end{tabular}
\setlength{\tabcolsep}{6pt}
}

  \caption{Test error on permutation invariant MNIST for different semi-supervised settings using the feedforward and the recurrent Neural Simpletrons with and without self-labeling.
We differentiate here between settings with different amounts of labels available during training.  
For results marked with $^{(*)}$, the free parameters were optimized using additional labels. We used the same parameter setting for all experiments shown here, which was tuned using 100~labels in total.
The results are given as the mean and standard error (SEM) over 100~independent repetitions, with randomly drawn, class-balanced labels.
In the fully labeled case, there are no unlabeled data points to use self-labeling on.
Therefore the results of ff- and ff\textsuperscript{+}-NeSi are identical there, as well as those of r- and r\textsuperscript{+}-NeSi.
}
  \label{tab:MNIST-results}
\end{table}

\begin{figure*}[p]
\begin{adjustbox}{trim=20pt 2pt 0 15pt}%
	\includegraphics{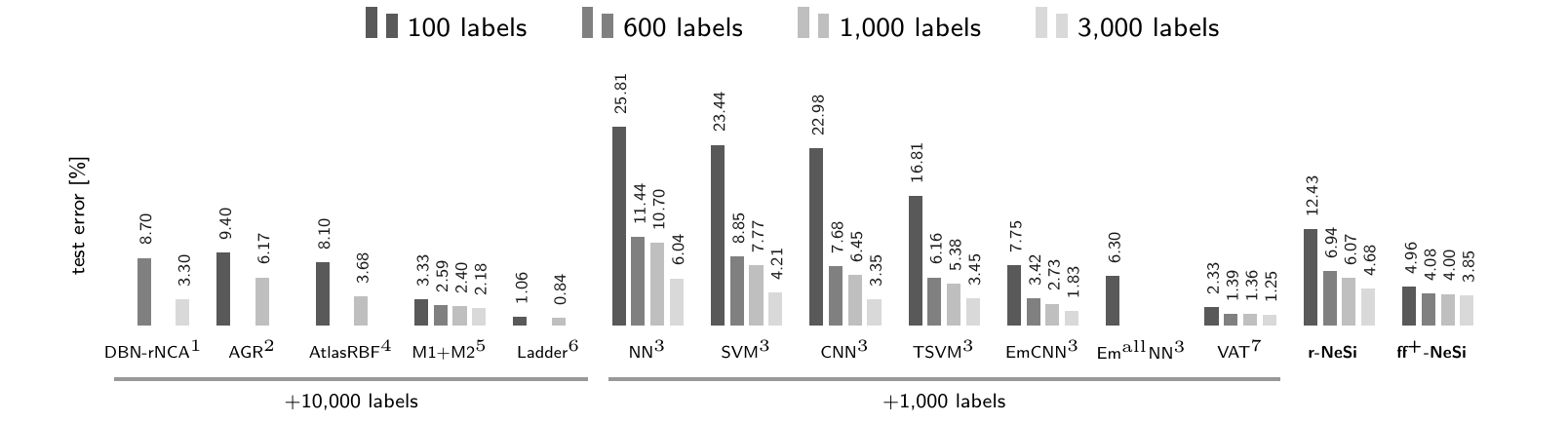}
\end{adjustbox}
\begin{adjustbox}{trim=20pt 0 0 0}%
	\includegraphics{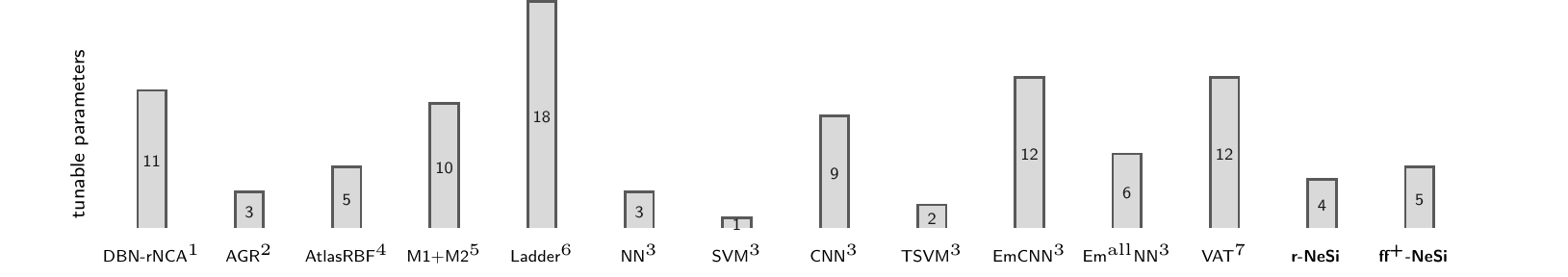}
\end{adjustbox}
\caption{Comparison of different algorithms on MNIST data with few labels.
The top figure shows results for systems using 100, 600, 1000, and 3000 labeled data points for training.
The algorithms are described in detail in the corresponding papers:
$^1${}\cite{SalakhutdinovHinton2007},
$^2${}\cite{LiuEtAl2010},
$^3${}\cite{WestonEtAl2012},
$^4${}\cite{PitelisEtAl2014},
$^5${}\cite{KingmaEtAl2014},
$^6${}\cite{RasmusEtAl2015},
$^7${}\cite{MiyatoEtAl2015}.
All algorithms except ours use \num{1000} or \num{10000}~additional data labels (from the training or test set) for parameter tuning.
The bottom figure gives the number of tunable parameters (as estimated in \cref{TabHyperParams}) and, where known, learned parameters of the algorithms (note the different scales).
}
\label{fig:MNIST-sota-Bar}
\end{figure*}

\Cref{fig:MNIST-sota-Bar} shows a comparison to standard and recent state-of-the-art approaches.
The NeSi networks are competitive---outperforming deep belief networks (`DBN-rNCA') and other recent approaches (like the `Embed'-networks, `AGR' and `AtlasRBF').
In the light of reduced model complexity and effectively used labels, we can furthermore compare to the few very recent algorithms with a lower error rate (`M1+M2', `VAT' and the `Ladder'-networks).

\begin{figure*}[p]
\centering
\begin{subfigure}{.3\textwidth}
	\centering
	\begin{adjustbox}{trim=0 0pt 0 0pt}%
		\includegraphics{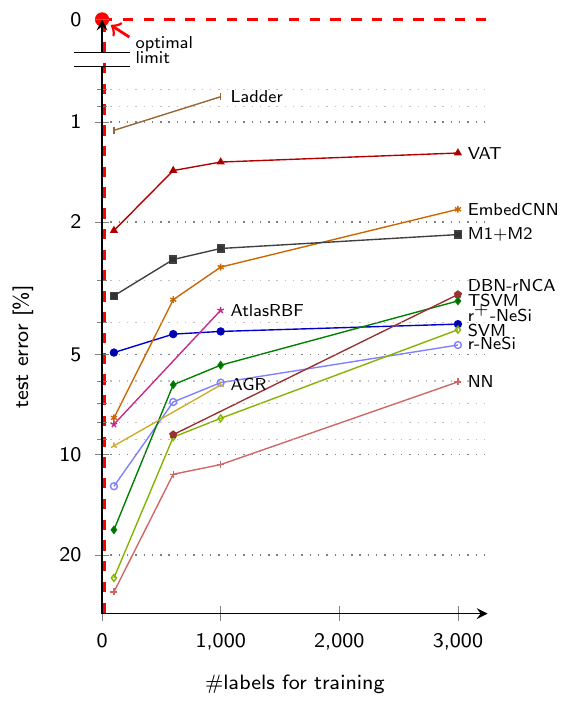}
	\end{adjustbox}
\end{subfigure}%
\begin{subfigure}{.7\textwidth}
  \centering
	\begin{adjustbox}{trim=0 0 40pt 0}%
		\includegraphics{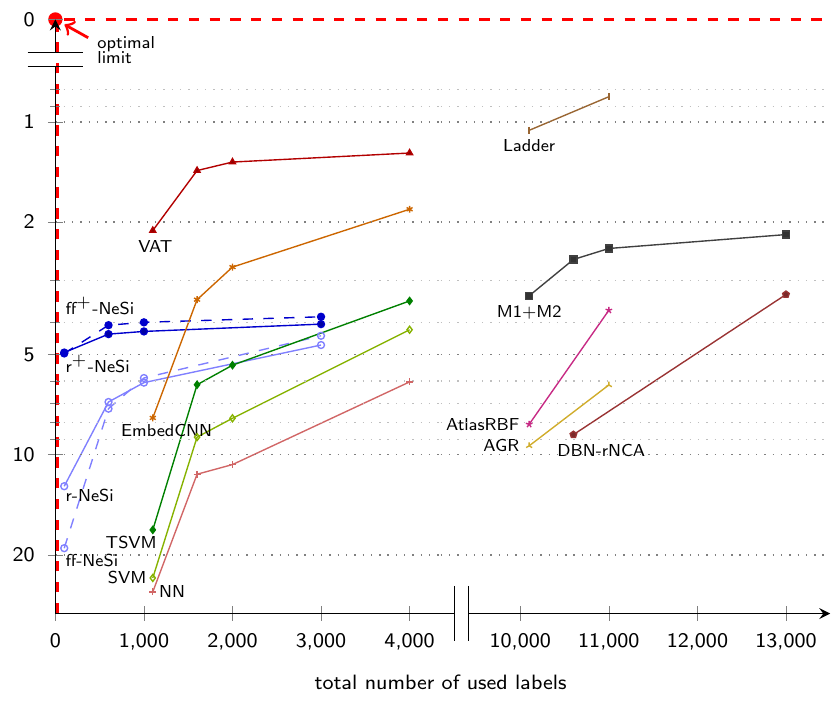}
	\end{adjustbox}
\end{subfigure}%
\caption{Classification performance of different algorithms compared against varying proportion of labeled training data.
The corresponding papers are listed in \cref{fig:MNIST-sota-Bar}.
The left-hand-side plot shows the achieved test errors w.r.t.\ the amount of labeled data seen by the compared algorithms during training.
The right-hand-side plot illustrates for the same experiments the total amount of labeled data seen by each of the algorithms over the whole tuning and training procedure.
For better readability, we only show the recurrent NeSi networks in the left-hand side plot.
Results for the feedforward networks can be directly transmitted from the right-hand side.
The plots can be read similar to ROC curves, in the way that the more a curve approaches the upper-left corner, the better is the performance of a system for decreasing amounts of available labeled data.
}
\label{fig:MNIST-sota-Scatter}
\end{figure*}

For the comparison in \cref{fig:MNIST-sota-Bar}, we have to point out, that (for lack of more comparable findings) all other algorithms are actually reporting results for a markedly differing (and generally easier) task than we do.
All of these models either use a validation set with a substantial amount of additional labels than available during training or the test set for parameter tuning.
Also, some of the algorithms (namely the TSVM, AGR, AtlasRBF and the Em-networks) actually train in the transductive setting, where the (unlabeled) test data is included into the training process.
For the NeSi approaches however, we avoided any training or tuning on the test set or on additional labeled data.
This also prevents the risk of overfitting to the test set.
The more complex a system is, the more labels are generally necessary to find optimal parameter settings that are not overfitted to a small validation set and generalize poorly.
When using test data during parameter tuning, the danger of such overfitting is even more severe as overfitting effects could be mistaken as good generalizability.
Therefore, in \cref{fig:MNIST-sota-Bar} we grouped the models by the amount of additional labeled data points used in the validation set for parameter tuning and also show the number of free parameters for each algorithm, as far as we were able to estimate from the corresponding papers.
These numbers have of course to be taken with high care, as not all parameters can be treated equally.
For some tunable parameters, for example, a default value may already always give good results, while others might have to be highly optimized for each new task.
Thus, these numbers should be taken more as an index for model complexity.

\Cref{fig:MNIST-sota-Scatter} shows the performance of the models with respect to the number of labels used during training (left-hand side) and with respect to the total number of labels used for the complete tuning and training procedure (right-hand side).
For the NeSi algorithms, these plots are identical, as we only use maximally as many labels in the tuning phase as in the training phase for the shown results.
For all other algorithms however, these plots can be regarded as the two extreme cases, where their actual performance in our chosen setting would probably lie somewhere in between.

\subsubsection{Overfitting Control for NeSi}
\label{sec:OverfittingControl}
With a network of \num{10000}~hidden units which learns on \num{60000}~training samples, some of the hidden units adapt to represent more rarely seen patterns while others adapt to represent patterns, that are more frequent in the training data.
Furthermore, the network learns the frequency at which patterns occur as the distribution $p(\subc|\R) = \frac{1}{\classK}\!\sum_\classk\! \Rkc$.
\Cref{fig:MNIST-weights} displays a random selection of 100 out of the \num{10000}~fields after training using the r\textsuperscript{+}-NeSi algorithm:
\begin{SCfigure}[][htb]
  \includegraphics[width=0.5\textwidth]{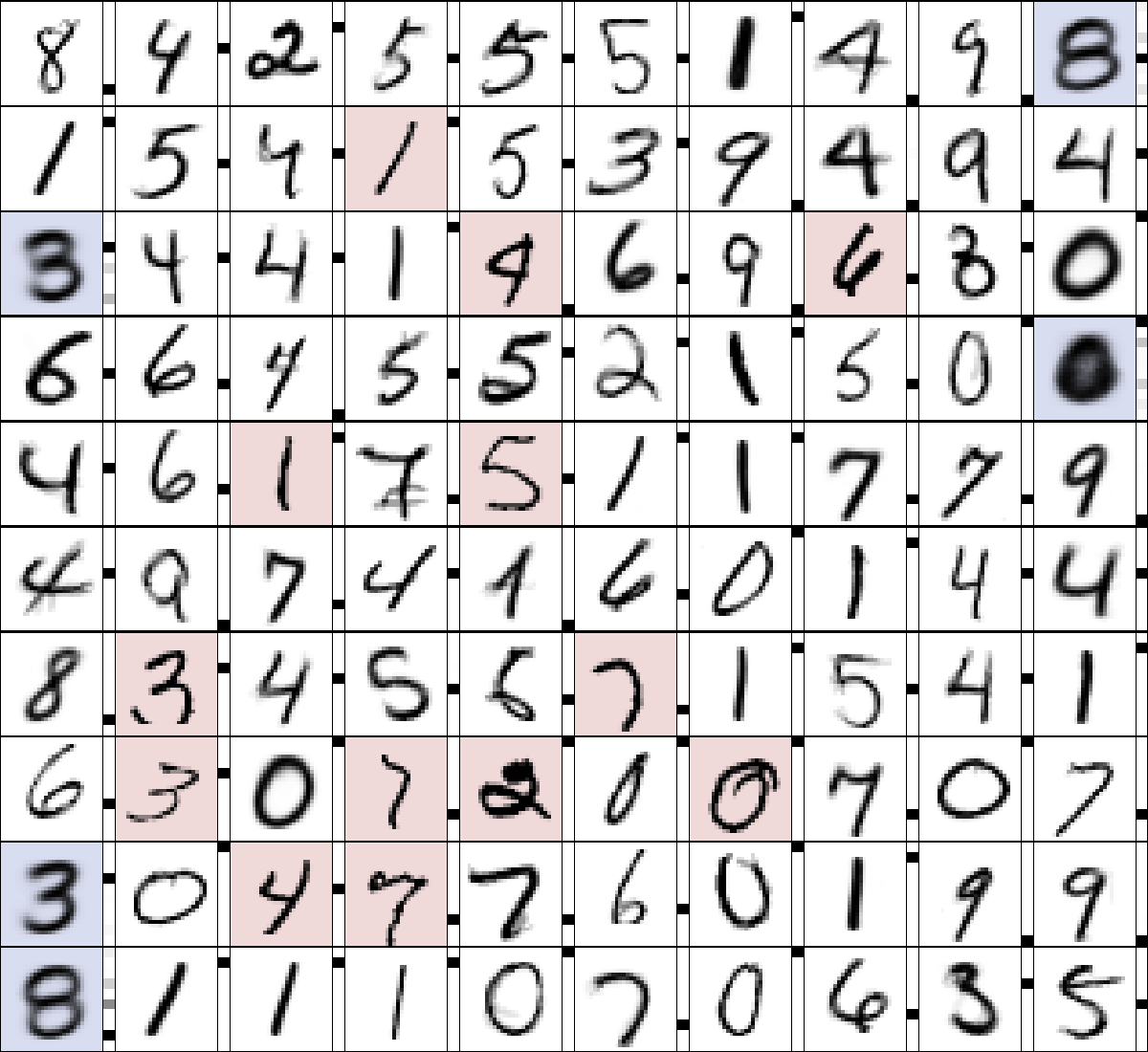}
  \caption{A subset of converged weights learned by the r\textsuperscript{+}-NeSi algorithm in the setting of 100~labels. The squared fields are 100 of the \num{10000} learned weights $\W[\subc][,:]$ with their learned class belonging $\R[:,][\subc]$ as columns next to the fields (starting with class `0' at the top to class `9' at the bottom of each column). Blue fields have \mbox{$p(\subc|\R)\cdot \inpN<0.5$}. Those are `forgotten fields' of the network whose connections are too weak for further specialization. The red fields have \mbox{$0.5\leq p(\subc|\R)\cdot\inpN<1.5$}. Those are fields that highly specialized to a single pattern in the training set.}
  \label{fig:MNIST-weights}
\end{SCfigure}

Fields colored in blue in \cref{fig:MNIST-weights} have a very low probability of $p(\subc|\R)\cdot\inpN<0.5$, with most of them $p(\subc|\R)$ being close to zero.
These fields have ceased to further specialize to respective pattern classes because of sufficiently many other fields that have optimized for a class.
They are effectively discarded by the network itself, as the low values in $\Rkc$ further suppress the activation of those fields in the recurrent network.
With longer training times, $p(\subc|\R)$ of those fields converges to zero, which practically prunes the network to the remaining size.
The red fields in \cref{fig:MNIST-weights} have a probability of $0.5 \leq p(\subc|\R)\cdot\inpN < 1.5$ to be activated, which corresponds to approximately one data point in the training set that activates the field. 
Such weights are often adapted to one single training data point with a very uncommon writing style (like the crooked `7' in 4th column, 9th row) or some kind of preprocessing artifact (like the cropped `3' in 2nd column, 7th row).

We did control for the effect of the rarely active fields (blue and red in \cref{fig:MNIST-weights}), especially as some of the fields are clearly overfitted to the training set. 
For that, we compared an original network of \num{10000} fields (that is, \num{10000} middle layer neurons) with a network for which all fields with activity $p(\subc|\R)\cdot\inpN < 1.5$ were removed (which was around 15\% of the \num{10000} fields).
We observed no significant changes in the test error between the original and the pruned network.
The reason is, that the pruned fields are essentially never activated at test time because of low similarities to test data and strong suppression by the network itself (due to the learned low activation rates during training).
\subsection{Large Scale Handwriting Recognition (NIST SD19)}
\label{sec:NIST}
Modern algorithms---especially in the field of semi-supervised learning---should be able to handle and benefit from the ever increasing amounts of available data (`big data').
A comparable task to MNIST, but with many more data points and much higher input dimensionality, is given by the NIST Special Database 19.
It contains over \num{800000} binary $128\times128$ images from \num{3600} different writers (with around half of the data being handwritten digits and the other half being lower and upper case letters). We perform experiments of both digit recognition (10 classes) and case-sensitive letter recognition (52 classes).
\begin{figure}[b]
  \includegraphics[width=\textwidth]{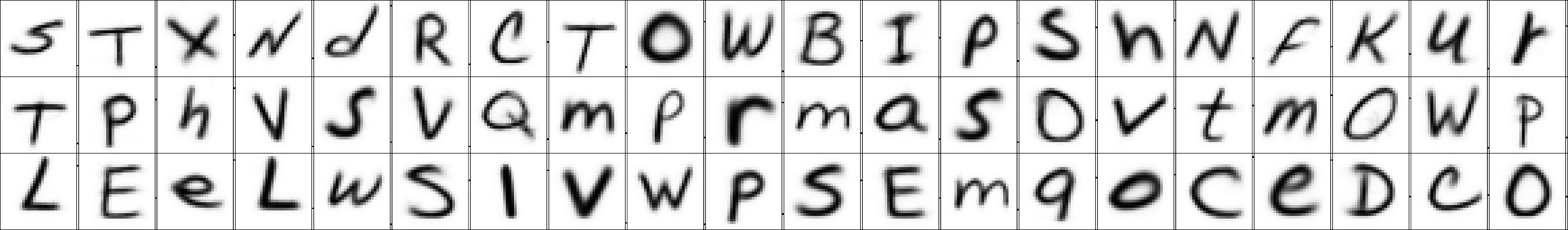}
  \caption{Visualization of learned weights of middle layer units of a ff$^+$-NeSi network when trained in the semi-supervised setting for NIST hand-written letters data using 520~labels (10 per class).
  }
  \label{fig:NIST-weights}
\end{figure}
We first applied the NeSi networks to the unpreprocessed NIST SD 19 digit data with $\dimD=\num{16384}$ input pixels.
The data is of much higher dimensionality than MNIST and the patterns are not centered by pixel mass, which represents a significantly more challenging task, as a lot more uninformative variation is kept within the data.
Hence, having a mixture model, learning these variations would need many more hidden units to achieve similar performance.
When keeping the same parameter setting as for MNIST (where we only increased $\normA$ to 25000, giving $\normA/\dimD \approx 1.5$, to account for the increased input dimensionality), the best performance for digit data in the fully labeled case was achieved by the r-NeSi network with an error rate of 9.5\%.

For better performance and easier comparison, we preprocessed the data similar to MNIST \citep[compare][]{CiresanEtAl2012}: for each image, we calculate square bounding boxes, resize to $20\times20$, zero-pad to $28\times28$ and center by pixel mass.
Finally, we invert the image, such that patterns have high pixel values instead of the background as is the case for MNIST.
For simplicity's sake and because of its high similarity, we then use the same setting for our free model parameters as we used for MNIST without further retuning.
The experiments are done using 1, 10, 60, 100, 300, or all labels per class.
We allowed for the same number of iterations as for MNIST to give sufficient training time for convergence.
However, with roughly five times more training data than for MNIST but the same total amount of labels, we now have a five times lower average activation in the top layer until self-labeling starts.
In the semi-supervised settings, we therefore scale the learning rate of the top layer also by a factor of five compared to MNIST to $\eR=1 \times \classK/\inpN$ for comparable convergence times.
\Cref{fig:NIST-weights} shows some examples of learned weights by the ff$^+$-NeSi network with 10~labels per class.
In \cref{tab:NIST-results}, we report the mean and standard error over 10~experiments on both digit and letter data.
For the NeSi networks, the results are given for the permutation invariant task.
To the best of our knowledge, this is the first system to report results for NIST~SD19 in the semi-supervised setting.

\begin{table}[t]
  \centering
  {
\fontsize{8}{10} \sffamily \sansmath
\renewcommand{\arraystretch}{1.3}
\newcommand{\midspace}{\hphantom{abc}}
\setlength{\tabcolsep}{9pt}
\scriptsize
\begin{tabular}{@{}l r@{}c@{}l r@{}c@{}l r@{}c@{}l r@{}c@{}l r@{}c@{}l r@{}c@{}l r@{}c@{}l}
\toprule
\#labels/class & \multicolumn{3}{c}{1} & \multicolumn{3}{c}{10} & \multicolumn{3}{c}{60} &  \multicolumn{3}{c}{100} & \multicolumn{3}{c}{300} & \multicolumn{3}{c}{fully labeled}\\
\midrule
\noalign{\vskip 2pt} 
\multicolumn{6}{@{}l}{digits (10 classes)}\\
\#labels total & \multicolumn{3}{c}{10} & \multicolumn{3}{c}{100} & \multicolumn{3}{c}{600} & \multicolumn{3}{c}{1000} & \multicolumn{3}{c}{3000} & \multicolumn{3}{c}{344\,307} \\
\midrule
ff$^+$-NeSi & \bfseries 7.56 & $\,\pm\,$ & 1.79 & \bfseries 6.20 & $\,\pm\,$ & 0.16 & 6.02 & $\,\pm\,$ & 0.08 & 6.02 & $\,\pm\,$ & 0.12 & \bfseries 5.70 & $\,\pm\,$ & 0.03 & 5.11 & $\,\pm\,$ & 0.01 \\
r$^+$-NeSi & 9.84 & $\,\pm\,$ & 2.40 & \bfseries 6.14 & $\,\pm\,$ & 0.23 & \bfseries 5.83 & $\,\pm\,$ & 0.14 & \bfseries 5.94 & $\,\pm\,$ & 0.12 & \bfseries 5.72 & $\,\pm\,$ & 0.10 & 4.52 & $\,\pm\,$ & 0.01 \\
35c-MCDNN &&&&&&&&&&&&&&&& \multicolumn{3}{c}{\bfseries 0.77}\\
\midrule
\noalign{\vskip 2pt} 
\multicolumn{6}{@{}l}{letters (52 classes)}\\
    \#labels total& \multicolumn{3}{c}{52} & \multicolumn{3}{c}{520} & \multicolumn{3}{c}{3120} &  \multicolumn{3}{c}{5200} & \multicolumn{3}{c}{15600} & \multicolumn{3}{c}{387361}\\
	\midrule
			ff$^+$-NeSi & \bfseries 55.70 & $\,\pm\,$ & 0.62 & \bfseries 46.22 & $\,\pm\,$ & 0.43 & 44.24 & $\,\pm\,$ & 0.23 & 43.69 & $\,\pm\,$ & 0.21 & 42.96 & $\,\pm\,$ & 0.28 & 34.66 & $\,\pm\,$ & 0.05 \\
    r$^+$-NeSi & 64.97& $\,\pm\,$ & 0.85 & 54.08 & $\,\pm\,$ & 0.38 & \bfseries 43.73 & $\,\pm\,$ & 0.15 & \bfseries 41.57 & $\,\pm\,$ & 0.13	 & \bfseries 37.95 & $\,\pm\,$ & 0.12	 & 31.93 & $\,\pm\,$ & 0.06 \\
    35c-MCDNN & & & & & & & & & & & & & & & & \multicolumn{3}{c}{\bfseries 21.01} \\
    \bottomrule
\end{tabular}
\setlength{\tabcolsep}{6pt}
}

  \caption{Test error on NIST SD19 data set on the task of digit and letter recognition for different total amounts of labeled data. The results for NeSi are permutation invariant and given as the mean and standard error (SEM) over 10 independent repetitions, with randomly drawn, class-balanced labels.}
  \label{tab:NIST-results}
\end{table}

Like for MNIST, the performance of our 3-layer network is in the fully labeled setting not competitive to state-of-the-art fully supervised algorithms \citep[like the 35c-MCDNN, a committee of 35~deep convolutional neural networks,][]{CiresanEtAl2012}. Note the difference, however, that our results do apply for the permutation invariant setting and do not take prior knowledge about two-dimensional image data into account (like convolutional networks). More importantly, for the settings with few labels, we only see a relatively mild decrease in test error when we strongly decrease the total number of used labels.
Even for just ten labels per class most patterns are correctly classified for the challenging task of case-sensitive letter classification (chance is below $2$\%).
Comparison of the digit classification setting with MNIST furthermore suggests, that not the relative but the absolute amount of labels per class is important for learning in our networks \citep[compare, for example,][footnote 4]{RasmusEtAl2015}.

In general, digit classification with NIST SD19 seems to be a more challenging task than MNIST \citep[which can also be observed in the results of][]{CiresanEtAl2012}.
However, the test error in our case increased slower than for MNIST with decreasing numbers of labels---and in the extreme case of a single label per class even surpassed the MNIST results.
When using, as for MNIST, only \num{60000} training examples for NIST, the test error for the single-label setting on digit data increased from $(7.56 \pm 1.79)\%$ to $(9.10 \pm 0.92)\%$ for ff$^+$-NeSi, showing the benefit of additional unlabeled data points.
The feedforward network with self-labeling is best in keeping a low test error with very few labels.
In fact, the main reason for the increase in test error for the single-label case are rare outliers, where two or more classes were learned completely switched, for example, all `3's were learned as `8's and vice versa.
This can happen, when the single randomly chosen labeled data points of two similar classes are too ambiguous and therefore lie close together at the border between two clusters.
This resulted in most networks within the 10~runs to have test errors between 5.5\% and 7\%, and one outlier at over 20\% (see \cref{AppResults}).
And it seems that additional unlabeled data points lead to better defined clusters, where this problem occurs less frequently.
Since in the recurrent network the label information is also fed back to the middle layer, this network is more sensible to label information.
On one hand, this helps when more label information is known.
On the other hand, this also more often results in a stronger accumulation of errors in the self-labeling procedure as wrong labels are less frequently corrected.

With more training data available than for MNIST, we also tried out bigger networks of \num{20000} hidden units for digit data, but only saw slight improvements on the test error.
This points to a limit of learnable subclasses (a.k.a.\ writing styles) within the data, where the modeling of more than $\subC = \num{10000}$ subclasses improves performance only very little but the increased amount of data in NIST helps to better define those given subclasses.

\subsection[Bio-Inspired Neural Networks]{Comparison to Bio-Inspired Neural Networks for Neuromorphic Hardware}\label{sec:BioInspired}

In addition to systems optimized for functional performance on standard CPU and GPU hardware, another line of research investigates learning systems that are well-suited for execution on alternative approaches such as analog VLSI circuits.
Most such systems are based on spiking neuron models and neurally plausible learning rules such as spike-timing-dependent-plasticity \citep[STDP;][]{GerstnerEtAl1996,BiPoo2001}.
A major advantage of learning algorithms implemented on analog VLSI chips are their time and energy efficiency compared to conventional hardware.
These features have the potential to make analog VLSI chips, which are in this context often referred to as neuromorphic chips, to a very high potential new hardware technology.

Architecture and task domain of bio-inspired learning systems share properties with deep neural networks and the simpletron systems discussed here, which makes comparison interesting.
We compare here to three recent versions of spiking neural networks that learn unsupervised on data. Notably, also for this research domain MNIST is used as a major tool for evaluation, which facilitates comparison.
We adapt the NeSi networks to relate more closely to the respective systems we compare to.
Except for the network size~$\subC$, we keep all free parameters at the optimized setting of \cref{sec:MNIST} and report test errors as the mean and standard error (SEM) over 10~independent training runs.

While bio-inspired systems are increasingly often realized on neuromorphic hardware \citep[for example,][]{SchmukerEtAl2014}, the results of the systems we compared to were obtained in simulations on conventional hardware as reported in the corresponding papers \citep[][]{DiehlCook2015,NeftciEtAl2015,NesslerEtAl2013}. 
Before we discuss comparison details, let us note, that the scope and goals of systems for neuromorphic hardware are different from those of the deep networks, our neural networks and the other systems discussed above.
A main goal being efficient implementability on neuromorphic hardware, which is not in the focus of deep learning systems.
Most bio-inspired systems for neuromorphic hardware are therefore based on spiking neurons as such neurons are routinely implemented on neuromorphic chips.
Neither our networks nor any of the other systems we considered above use spiking neurons.

Let us first consider the bio-inspired spiking neural network (SNN) model of \citet[][]{DiehlCook2015}, which consists of an input layer and a single hidden layer of up to \num{6400}~spiking neurons.
Results for MNIST are obtained after transforming the observed data to Poisson spike trains.
After training, a class label is assigned to each field by determining their highest average activation per class on the training data.
Such a training procedure is comparable to the ff-NeSi algorithm, which also learns the first hidden layer completely unsupervised and only uses data labels to assign classes to the learned representations in a separate training stage.
However, instead of using a max-assignment as in \citet{DiehlCook2015}, we use an additional neural layer which approximates a Bayesian classifier (Eqs.~T1.6 and T1.8) and learns the complete conditional distribution $p(\subc|\classk,\R) = \Rkc$.
When scaled to \num{6400}~neurons, the network of \citet[][]{DiehlCook2015} achieves a $5.0\%$~error rate on MNIST.
With our similar (but non-spiking) ff-NeSi network of \num{6400}~neurons, we obtained a test error of $(3.28 \pm 0.04)\%$.
To make the systems still more similar, we used the same max assignment of top-layer weights by \citet[][]{DiehlCook2015} also for our ff-NeSi network, that is, we assigned to each field the single unweighted label which corresponds to the class that activated the field most in the training set.
When using this hard assignment, we found that our test error increased from $(3.28 \pm 0.04)\%$ to $(3.62 \pm 0.03)\%$ for the fully labeled case, showing a benefit of a
probabilistic treatment.
If we, like before, only used 100~random class balanced labeled data points for the class assignment of fields, we achieved a classification error of $(19.97 \pm 0.84)\%$ for the ff-NeSi network with \num{6400}~neurons, and $(21.00 \pm 0.86)\%$ when we used the hard class assignment of \citet{DiehlCook2015}.
Using self-labeling instead, ff\textsuperscript{+}-NeSi achieved a test error of $(5.10 \pm 0.18
)\%$.
The semi-supervised settings have not been investigated by \citet{DiehlCook2015} but it would represent interesting data for comparison, and it should be straightforward to operate the spiking network model also in this regime.

The second system we compare to is the Synaptic Sampling Machine (SSM), recently suggested by \citet[][]{NeftciEtAl2015}. 
The network consist of $28\times{}28=784$~neurons in the input layer, $500$~neurons in the first hidden layer and $10$~neurons in the top layer.
Inference and learning is implemented based on spiking neurons with MNIST data represented by Poisson spike trains.
The SSM is closely related to Restricted Boltzmann Machines \citep[RBMs; see, for example,][]{DayanAbbott2001,SalakhutdinovHinton2009} with weight changes following a continuous time variant of contrastive divergence \citep[][]{Hinton2002}.
If trained on the MNIST training set using all labels, the SSM achieves an error rate of $4.4\%$ on the test set.
In addition to the SSM, \citet[][]{NeftciEtAl2015} also considered standard discrete time RBMs with the same architecture (784--500--10).
Using the most conventional setting with Gibbs sampling and standard contrastive divergence, the RBM obtained an optimal test error of $5.0\%$ (again using all labels).
Learning for the RBM was here assumed to stop at the point of optimal performance while longer learning resulted in larger test errors which were attributed to decreased MCMC ergodicity and overfitting \citep[][]{NeftciEtAl2015}.
An improved RBM variant, the dSSM network, did not suffer from such overtraining effects.
The test error of the dSSM (architecture 784--500--10) on fully labeled MNIST was 4.5\%.
The SSM, RBM and dSSM systems have essentially the same network architecture as our NeSi systems if we use $500$~middle layer neurons.
For this setting (without further optimization of the remaining free parameters), the ff-NeSi network achieved a test error of $(4.95 \pm 0.03)\%$ and r-NeSi of $(3.97 \pm 0.03)\%$ (both fully labeled).
When using only \num{100}~labels for training, r\textsuperscript{+}-NeSi achieved a test error of $(11.50 \pm 1.37)\%$ and ff\textsuperscript{+}-NeSi of $(8.36 \pm 0.64)\%$.

So far, we used the usual training setup, where classes are learned from labeled training data.
An alternative evaluation procedure is suggested by \citet{NesslerEtAl2013} for the final system we compare to.
The spike-based Expectation Maximization approach (SEM) implements a generative Poisson model as spiking neural network with STDP rules.
The network is trained unsupervised with \num{100}~neurons on MNIST.
The class assignment of the learned representations is then done directly by the user who inspects the fields and assigns to each field what he considers the most likely label.
With this procedure, the network of \citet{NesslerEtAl2013} achieves a test error of $19.86\%$.
We can adopt the same procedure by only training the first layer of an ff-NeSi network and then assigning the fields manually with labels~$\labely^{(\subc)}$ by setting the weights $\Rkc$ to 
\begin{align}
\Rkc = \frac{\delta_{\labely^{(\subc)}\classk}}{\sum_{\subc'} \delta_{\labely^{(\subc')}\classk}}.
\end{align}
Using this procedure, the NeSi network achieved a test error of $(10.53 \pm 0.11)\%$. 
For functional goals, we could further improve on these results.
We could, for example, ask the user to assign a certainty weight to the chosen labels or even ask to assign a probability distribution over all possible labels.
Improvements are also possible without requesting further information from the supervisor.
By using recurrence and self-labeling of the r\textsuperscript{+}-NeSi network, we were able to improve classification down to an error rate of $(5.15 \pm 0.26)\%$ based on 100~fields labeled by the user (see \cref{sec:InteractiveLabeling} for details).

Notably, other lines of research \citep[for example,][]{EsserEtAl2015,DiehlEtAl2015} do not consider networks for spike-based
learning and inference but focus on spike-based inference alone. Typically, in a first stage, standard (non-spiking) discriminative networks are trained using conventional back-propagation, and only afterwards, in
a second stage, the trained networks are translated to spiking versions that can be implemented on neuromorphic hardware. As such approaches are, in this sense, no spike-based learning systems, and because of their fully supervised setting inherited from standard deep learning, they are not considered here as spike-based learning networks.

A summary of the most relevant comparison results is given in \cref{tab:BioResults}.
\begin{table}[!ht]
  \centering
  {
\fontsize{8}{10} \sffamily \sansmath
\renewcommand{\arraystretch}{1.3}
\setlength{\tabcolsep}{8pt}
\scriptsize
\begin{tabular}{@{}l c c r@{\,}c@{\,}l}
\toprule
algorithm & \hspace{-12pt} 2nd layer neurons & class assignment & \multicolumn{3}{c}{test error $[\%]$} \\
\midrule
SNN \citep{DiehlCook2015} & 6400 & hard max & \multicolumn{3}{c}{$5.0$} \\
ff-NeSi & 6400 & hard max & $3.62$ & $\pm$ & $0.03$ \\
ff-NeSi & 6400 & implicit & $3.28$ & $\pm$ & $0.04$ \\
\midrule
SSM \citep{NeftciEtAl2015} & 500 & implicit & \multicolumn{3}{c}{$4.4$} \\
dSSM \citep{NeftciEtAl2015} & 500 & implicit & \multicolumn{3}{c}{$4.5$} \\
RBM \citep{NeftciEtAl2015} & 500 & implicit & \multicolumn{3}{c}{$5.0$} \\
ff-NeSi & 500 & implicit & $4.95$ & $\pm$ & $0.03$ \\
r-NeSi & 500 & implicit & $3.97$ & $\pm$ & $0.03$ \\
\midrule
SEM \citep{NesslerEtAl2013} & 100 & user & \multicolumn{3}{c}{$19.86$} \\
ff-NeSi & 100 & user & $10.53$ & $\pm$ & $0.11$ \\
\bottomrule
\end{tabular}
\setlength{\tabcolsep}{6pt}
}

  \caption{Comparison with bio-inspired systems on MNIST. SNN \citep{DiehlCook2015}, SSM, dSSM \citep{NeftciEtAl2015} and SEM \citep{NesslerEtAl2013} are spiking neural networks, while the NeSi networks are non-spiking.
  The systems are sorted by the number of neurons in the 2nd layer (first hidden layer) and the blocks separated by horizontal lines group those systems that are the most similar to each other.
  The standard NeSi setup is explicitly changed to facilitate comparison: `hard max' refers to the class assignment of \cite{DiehlCook2015} and `implicit' refers to the (different) standard procedures to assign class labels for SSMs, RBMs, or the NeSi systems.
  In both cases learning used all labels of the MNIST training set (note that the ff\textsuperscript{+}- and r\textsuperscript{+}-NeSi versions are irrelevant for this setting).
  For class assignment `user', learned fields were hand-labeled and no labels of the training set were used.
  All test errors were computed for the MNIST test set.
  We show the mean and standard error (SEM) based on 10~runs for the NeSi systems.
  Other values were taken from the respective publications.}
  \label{tab:BioResults}
\end{table}

\section{Discussion}
Deep learning is an important and highly successful research field with approaches filling a spectrum of algorithms from purely feedforward and discriminative neural networks to directed generative models.
Deep discriminative neural networks (DNNs) dominate the field, especially in the prominent domain of classification tasks.
By deriving the NeSi algorithms from a directed generative model, we have shown in this study that inference and learning in a deep directed graphical model can take a very similar form as learning in standard DNNs.
Furthermore, the derived networks, which we called {\em Neural Simpletrons} (NeSi), do in our empirical comparison improve on all standard deep neural networks (like deep belief networks and CNNs) when only limited amounts of labeled data are available, and they are competitive to very recent deep learning approaches.

{\em Relation to Standard and Recent Deep Learning.}
Neural Simpletrons are, on the one hand, similar to standard DNNs as they learn online (that is, they learn per data point or per mini-batch), are efficiently scalable, and as their activation and learning rules are local, elementary, and neurally plausible (see \cref{tab:learningrules}).
On the other hand, the NeSi networks exhibit features that are a hallmark of deep directed generative models such as learning from unlabeled data and integration of bottom-up and top-down information for optimal inference. 
By comparing the learning and neural interaction equations of DNNs and the NeSi networks directly, Eq.~(T1.5) for top-down integration and the learning rules Eqs.~(T1.7) and (T1.8) represent the crucial differences.
The first one allows the NeSi networks to integrate top-down and bottom-up information for inference, which contrasts with pure feedforward processing in DNNs.
The second one shows, that NeSi learning is local and Hebbian while approximating likelihood optimization, which contrasts with less local back-propagation for discriminative learning in standard DNNs. 
In the example of the NeSi networks, recurrent bottom-up/top-down integration was especially useful in the fully labeled case (particularly in the complete setting, see \cref{sec:20NewsFullyLabeled}).
When we acquire additional inferred labels through self-labeling, the feed-forward system was best in maintaining a low test error even down to the limit of a single label per class.
For fully labeled data, the NeSi systems are not competitive anymore, as seen, for example, on MNIST.
Discriminative approaches dominate in this regime as it seems to be difficult to compete with discriminative learning with such a minimalistic system once sufficiently many labeled data points are available.
Furthermore, the generative NeSi approach relies on the possibility to learn representations of meaningful templates
(as shown, for example, in \cref{fig:20News-weights,fig:MNIST-weights,fig:NIST-weights}); and template representations make the networks very interpretable.
However, for example for large image databases showing 2-D images of 3-D objects, learning of such templates based on pixel intensities seems very challenging.
For NeSi networks, an additional feature layer (with additional parameters) is likely to facilitate the learning of template representations.
Such a requirement would however significantly divert from our study of a minimalistic system.

Besides of the approaches studied here, many other systems are able to make use of top-down and bottom-up integration for learning and inference.
Top-down information is provided in an indirect way if a system introduces new labels itself by using its own inference mechanism.
Similar to the ff\textsuperscript{+}- and r\textsuperscript{+}-NeSi networks, this self-labeling idea has been followed repeatedly previously \citep[for a recent overview, see][]{TrigueroEtAl2015}.
For the NeSi systems, such feedback worked especially well, which may indicate that self-labeling is particularly promising for deep directed models.
Systems that make a more direct use of bottom-up and top-down information include approaches based on undirected graphical models.
The most prominent examples, especially in the context of deep learning, are deep restricted Boltzmann machines (RBMs).
While RBMs are successfully used in many contexts \citep[for example,][]{HintonEtAl2006,GoodfellowEtAl2013,NeftciEtAl2015}, performance of RBMs alone, without
hybrid learning approaches, does not seem to be competitive with recent results on semi-supervised learning.
The best performing RBM-related systems we compared to here, are the HDRBM \citep[][]{LarochelleBengio2008} for 20~Newsgroups and the DBN-rNCA system \citep[][]{SalakhutdinovHinton2007} for MNIST.
Both approaches use additional mechanisms for semi-supervised classification, which can be taken as evidence for standard RBM approaches being more limited when labeled data is sparse.
In this semi-supervised setting, both ff-NeSi and r-NeSi perform better than the DBN-rNCA approach for MNIST (\cref{fig:MNIST-sota-Bar,fig:MNIST-sota-Scatter}) and better than the HDRBM for 20~Newsgroups (\cref{tab:20News-results}).
When optimized for the fully labeled setting, NeSi even improves considerably to the HDRBM in the fully labeled 20~Newsgroups task.
Recent RBM versions, enhanced and combined with discriminative deep networks \citep[][]{GoodfellowEtAl2013}, outperform NeSi networks on fully labeled MNIST---however, competitiveness in semi-supervised settings has not been shown, so far.
In our empirical evaluations, we also compared to a non-hybrid RBM approach more directly.
When using the very same network architecture (same layer and same neuron numbers), ff-NeSi and r-NeSi performed better than the RBM for fully labeled MNIST (see comparison to bio-inspired systems below).

Other approaches that can make use of bottom-up and top-down information are algorithms based on other types of directed graphical models.
Inference in such approaches is naturally probabilistic, recurrent, and of high interest from the functional and biological perspectives \citep[see, for example,][]{LeeMumford2003,HaefnerEtAl2015}.
Regarding the learning and inference equations themselves, the compactness of the equations defining the NeSi algorithms and their formulation as minimalistic neural networks represent a major difference to pure generative approaches \citep[such as][]{SaulEtAl1996,LarochelleMurray2011,GanEtAl2015} or combinations of DNNs and graphical models \citep[for example,][]{KingmaEtAl2014}.
Regarding empirical comparisons, typical directed generative models are not compared on typical DNN tasks but use other evaluation criteria.
Prominent or recent examples such as deep SBNs \citep[see, for example,][]{SaulEtAl1996,GanEtAl2015} have, for instance, not been shown to be competitive with standard discriminative deep networks on semi-supervised classification tasks, so far.
In general, a main challenge is the necessity to introduce approximation schemes.
The accuracy of approximations for large networks, and the complexity of the networks themselves, still seem to prevent scalability and/or competitive performance on tasks as discussed here.
In principle however, deep directed generative models such as deep SBNs or other deep directed multiple-cause approaches are more expressive than deep mixture models.
We thus interpret our results as highlighting the general potential of deep directed generative models also for tasks such as classification.

{\em Relation to Bio-Inspired Systems.}
Deep neural networks owe much of their success to their efficient implementation on standard hardware such as CPUs and, more so, state-of-the-art GPUs.
Another line of research focuses on non-standard hardware such as neuromorphic chips.
A primary goal in that field is the implementability of learning algorithms using spiking neurons, since most of neuromorphic developments use these as elementary building blocks.
Many of such bio-inspired systems are similar to deep learning systems or other classifier approaches.
The SSM system suggested by \citet[][]{NeftciEtAl2015} is for instance closely related to RBMs \citep[DBN;][]{HintonEtAl2006} and the SEM approach by \citet[][]{NesslerEtAl2013} uses EM to derive the spiking neural network.

In comparison to the NeSi approaches considered in this study, the SSM system and related RBM approaches \citep[see][]{NeftciEtAl2015} are the most similar bio-inspired approaches in terms of the network architecture.
Furthermore, r-NeSi, SSMs and RBMs are all able to integrate bottom-up and top-down information for inference.
Their respective inference and learning equations are different, however:
SSM and RBMs are based on undirected graphical models, use sampling procedures for inference and a contrastive divergence variant for learning.
In contrast, the NeSi networks are derived from a directed graphical model and use inference and learning equations as online approximations of exact EM.
Functionally, SSM, RBM and NeSi approaches achieve similar results in the setting investigated
by \citet[][]{NeftciEtAl2015} (small network architecture of 784--500--10).
The similar performance for the fully labeled case seems to highlight the common generative model nature of SSMs, RBMs and NeSi approaches.
The r-NeSi approach has with $3.97\%$ the lowest error rate compared to SSMs, RBMs and other bio-inspired systems.
For this comparison, $3.97\%$ is a low error if we consider that the SSM is with $4.4\%$ the currently best performing network with spike-based learning.
Recurrent inference used by r-NeSi, that is, the ability to integrate bottom-up and top-down information, is important for such a high performance.
Still, also the ff-NeSi system without recurrent inference achieves $4.95\%$, which is slightly lower than the error rate of a standard recurrent RBM. 

In contrast to SSM, RBMs and NeSi, the SEM approach by \citet[][]{NesslerEtAl2013} and related networks \citep[for example,][]{NesslerEtAl2009} use a shallow network architecture (one input and one hidden layer).
In terms of inference and learning between the neurons, the SEM network is however the system most closely related to the NeSi approaches.
Both, SEM and NeSi use a Poisson noise model for the bottom layer and learning is in both cases derived using the EM algorithm.
While Poisson noise is a well-suited distribution for positive observables, it also results in inference equations (E-step) with weighted activity summation and softmax-like lateral competition among hidden units (due to explaining away effects).  
Such properties enable neural network formulations of learning and inference as was shown, for example, by \citet{LuckeSahani2007b}, \citet{LuckeSahani2008}, \citet{KeckEtAl2012} and \citet{NesslerEtAl2013}, and for related distributions, for example, by \citet{DeneveEtAl2008} and \citet{NesslerEtAl2009}. 
By combining Poisson noise of a mixture model with explicit normalization, the network by \citet[][]{KeckEtAl2012} and the hierarchical NeSi networks arrive at very compact rate-based neural update and learning equations.
Following the goal of realizing spike-based networks, \citet[][]{NesslerEtAl2013} show how EM based learning for Poisson mixtures can be approximated using STDP.
Functionally, SEMs can learn unsupervised without using any labels.
If the learned fields are assigned to digit classes by the user, test error rates can be computed for the SEM \citep[][]{NesslerEtAl2013}.
By applying the same procedure to the ff-NeSi network, the SEM and NeSi systems can be compared: using 100~first hidden layer units in both networks, SEM achieves a test error of $19.86\%$ vs.\ $10.53\%$ for the ff-NeSi system. 

For SNN, SSM, SEM and other spike-based learning networks, performance for scales larger than those compared in \cref{tab:BioResults} would be interesting to investigate.
In general, such large spiking networks remain a challenge, however.
On the one hand, the number of neurons and the number of synapses that can be implemented on current neuromorphic chips is still relatively limited.
On the other hand, simulations of spiking networks on standard hardware require the simulation of the neural spiking dynamics, which represents a considerable overhead in computational effort.
The results for the scalable but non-spiking NeSi systems may therefore be taken as evidence in favor of the generally possible performance achievable by large scale spiking neural networks.

{\em Empirical Performance, Model Complexity and Data with Few Labels.}
The main focus in our study has been the semi-supervised task, especially in the limit of few labels.
Such a regime should here not be considered as a special boundary case.
Much to the contrary, with increasing capabilities of state-of-the-art sensors and data collected through other sources, large data sets are and will be increasingly easy to obtain.
Data labels are, on the other hand, costly and often erratic.
The limit of few labels is therefore arguably the most natural setting for many new applications \citep[also see discussions in][]{CollobertEtAl2006,KingmaEtAl2014}.

Our main results for the NeSi systems were obtained using the 20~Newsgroups, the MNIST and the NIST SD19 data sets (with MNIST simply being the data set for which most empirical data for semi-supervised learning is available).
\Cref{tab:20News-results,tab:MNIST-results,fig:MNIST-sota-Bar,fig:MNIST-sota-Scatter} summarize the empirical results and those used for comparison.
The r-NeSi system is the best performing system for the semi-supervised 20~Newsgroups data set, but the data set is much more popular as a fully supervised benchmark (comparison only to HDRBM in the semi-supervised setting).
More instructive for comparison is therefore the semi-supervised MNIST benchmark.
As can be observed in \cref{fig:MNIST-sota-Bar}, the NeSi algorithms achieved smaller test errors than all standard deep learning and a number of very recent classifier approaches.
Even hybrid systems enhanced for semi-supervised learning such as the DBN-rNCA approach or the EmbedCNN perform less well than, for example, the  r\textsuperscript{+}- and ff\textsuperscript{+}-NeSi algorithms.
Only three very recent approaches, M1+M2 \citep[][]{KingmaEtAl2014}, VAT \citep[][]{MiyatoEtAl2015}, and the Ladder network \citep[][]{RasmusEtAl2015} show a smaller test error than the NeSi approaches for data with few labels.
However, all of these systems are hybrid approaches: M1+M2 \citep[][]{KingmaEtAl2014} combines generative and back-prop learning approaches; the results for the VAT \citep[][]{MiyatoEtAl2015} are obtained by combining a DNN using back-prop with a smoothness constraint derived from the data distribution; and the ladder network \citep[][]{RasmusEtAl2015} applies a per-layer denoising objective onto standard discriminative learning models like MLPs and CNNs.
M1+M2 hereby shares the use of generative models with NeSi networks, and both approaches can be taken as evidence for two hidden layers of generative latents already resulting in competitive performances.
A difference is, however, the strong reliance of M1+M2 on deep neural networks to parameterize the dependencies between observed and hidden variables and dependencies among hidden variables which are optimized using DNN gradient approaches (the same applies for DNNs used for the applied variational approximation).
Inference and learning in M1+M2 is therefore significantly more intricate, and requires multiple deep networks.
Also the generative description part itself is very different (for instance, motivated by easy differentiability based on continuous latents) and is in M1+M2 not directly used for inference.
For Neural Simpletrons, the generative and the neural network connections are identical, and are directly used for inference.
Compared to all strongly performing recent approaches \citep[for example,][]{KingmaEtAl2014,MiyatoEtAl2015,RasmusEtAl2015}, the NeSi networks could, therefore, be considered as the best performing non-hybrid approach in terms of the numerical comparisons in \cref{sec:NumericalExperiments}.

If we compare the considered (hybrid and non-hybrid) systems in more detail (see \cref{fig:MNIST-sota-Bar}) a performance vs.\ model complexity trade-off can be observed.
If we consider the learning and tuning protocols that were used for the different systems to achieve the reported performance, large differences in the number of tunable parameters, the size of validation sets and the complexity of the systems can be noticed.
While some systems only need to tune few parameters, others (especially hybrid systems) require tuning of many free parameters (\cref{fig:MNIST-sota-Bar}).
Parameter tuning can be considered as a second optimization loop requiring labels additionally to those of the training set.
These additional labels (usually those of the validation set) are typically not taken into account if performance on semi-supervised settings are compared.
Some models use up to \num{10000} additional labels to tune their free parameters.
To (partly) normalize for model complexity, performance comparison w.r.t.\ the total number of required labels could therefore serve as a kind of empirical Occam's Razor. 
If this total number of labels is considered, the comparison between system performance changes as illustrated in \cref{fig:MNIST-sota-Scatter} (right-hand-side plot).
Considering \cref{fig:MNIST-sota-Scatter}, the VAT system (\num{1000} additional labels) could be considered to perform more strongly than the Ladder network if compared on the total number of labels.
The figure also shows that no other system has been shown to operate in a regime of as few labels as were used by the NeSi systems.
Especially when using self-labeling, NeSi networks can be applied to as few as 100~labels for the complete tuning and training procedure, where the algorithm achieves less than 5\% error on the MNIST test set.
While the numbers of tunable parameters for the different systems and the sizes of the used validation sets are clearly correlated (\cref{fig:MNIST-sota-Bar}), it remains unclear how many additional labels would really be required by the different systems.
The two plots of \cref{fig:MNIST-sota-Scatter} could therefore be considered as two limit cases for comparison.
\enlargethispage{\baselineskip}

{\em Future Work and Outlook.} 
As the NeSi networks share many properties with standard deep neural networks, further enhancements such as network pruning, annealing or drop-out could be investigated to further increase performance or efficiency.
Any new technique would make the algorithms more complex and introduce new parameters, however, which would take us further away from our goal of a minimalistic generative neural network.
The same would apply for any additional neural layer.
Still, future extensions could consider more than three layers (more than one middle layer) or layers with multiple separated softmax functions, which would lead into a committee-like approach.
Preliminary experiments with five separated softmax functions over \num{10000} middle layer hidden units each ($C=\num{50000}$ in total) with r$^+$-NeSi on MNIST already showed improvements of the test error, for example, from ($4.93 \pm 0.05$)\% to ($4.53 \pm 0.07$)\% when using 100~labels.
Also, the combination with discriminative learning approaches is a promising extension.
Ideally, such a combination would maintain a monolithic architecture and a limited complexity.
Other studies have already shown, that deep discriminative models can be related to directed generative models in grounded mathematical ways \citep[see][for a recent example]{PatelEtAl2016}.
Similarly, complementary discriminative methods could be derived for the NeSi systems.
Alternatively, co-training setups with more loosely coupled discriminative and generative learning can be investigated.
Other future extensions of the NeSi systems may involve generalizations to other types of input.
Input layers with other distributions including Gaussian noise could be investigated (such that also observables with negative values can be processed).
On the other hand, the Poisson distribution is well suited to process data that signal the presence and absence of features (describing sums of Bernoulli distributed observables).
This would motivate the use of a NeSi approach in combination with dictionaries learned with binary latents \citep[][]{LuckeSahani2008,GoodfellowEtAl2012,MohamedEtAl2012,SheikhEtAl2014}.
Further research directions would be combinations with hyperparameter optimization approaches \citep[for example,][]{ThorntonEtAl2013,BergstraEtAl2013,HutterEtAl2015} in order to increase autonomy and to exploit the very low number of free parameters.
Finally, the probabilistic nature of the NeSi networks would allow to address problems such as label noise in straightforward ways, while its generative model relation would allow for the investigation of tasks other than classification.

\clearpage
\appendix
\crefalias{section}{appendix}
\crefalias{subsection}{appendix}
\counterwithin{figure}{section}
\counterwithin{table}{section}

\belowdisplayskip=5pt
\abovedisplayskip=5pt

%
%
\section{Derivation Details}
\label{sec:AppModelDerivation}
Although the resulting NeSi neural network models exist as a very compact and simple set of equations, shown in \cref{tab:learningrules}, the derivation of these equations is not necessarily trivial.
Therefore, we give here further insight into some derivation steps to allow for a better understanding of the model at hand.
In \cref{sec:AppEMDetails}, we give details on the derivation of the EM update rules for the underlying generative model.
In \cref{SecOnlineLearning} we show the necessary derivation steps to attain the approximate equivalence of neural online learning with EM batch learning at convergence, which is the basis of our neural network derivation.

\subsection{EM Update Steps}
\label{sec:AppEMDetails}
\paragraph{E-Step.}
The posterior $p(\classk|\subc,\labely,\Theta)$ can be easily obtained by simply applying Bayes' rule for the labeled and unlabeled case.
For $p(\subc|\yVec,\labely,\Theta)$ however, some additional steps are necessary to attain the compact form shown in \cref{eq:PostC}:

We start again with Bayes' rule and use the sum and product rule of probability to regain the conditionals \cref{eq:PriorK,eq:PriorC,eq:Obs} of the generative model:
\begin{align}
p(\subc|\yVec,\labely,\Theta) 
	&= \frac{p(\yVec|\subc,\Wgen)\sum_\classk p(\subc|\classk, \Rgen)p(\classk|\labely)}
	{\sum_{\subc'} p(\yVec|\subc',\Wgen)\sum_\classk p(\subc'|\classk, \Rgen)p(\classk|\labely)}. \label{eq:EStep1} \\
%
\intertext{When we now insert the corresponding distributions \cref{eq:PriorC,eq:Obs} into \cref{eq:EStep1}, the benefit of assuming Poisson noise for $p(\yVec|\subc,\Wgen)$ becomes apparent: First, the factorial given by the $\Gamma$-function directly drops out. Second, by using the weight constraint \cref{eq:Obs}, the product of exponentials $\prod_{\dimd}e^{-\Wgencd} = e^{-\sum_\dimd\Wgencd}=e^{-\normA}$ also cancels with the denominator:}
	\dots & = \frac{\prod_{\dimd}\!\big((\Wgencd)^{\yd}\Gamma (\ydN \! + \! 1)^{-1}\, e^{-\Wgencd} \big) \sum_\classk \Rgenkc \uk}
	{\sum_{\subc'}\! \prod_{\dimd}\! \big( (\Wgen[\subc'][\dimd])^{\yd}\Gamma (\yd \! + \! 1)^{-1}\, e^{-\Wgen[\subc'][\dimd]}\big) \sum_\classk \Rgen[\classk][\subc'] \uk} \nonumber \\[2pt]
	&= \frac{\prod_{\dimd}(\Wgencd)^{\yd} \sum_\classk \Rgenkc \uk}
	{\sum_{\subc'}\! \prod_{\dimd} (\Wgen[\subc'][\dimd])^{\yd} \sum_\classk \Rgen[\classk][\subc'] \uk}\text{, with}\\[6pt]
& \uk = \left\{ \begin{array}{ll}p(\classk|\labely)=\delta_{\classk\labely} & \textnormal{\footnotesize for labeled data}\\ p(\classk)=\frac{1}{\classK} & \textnormal{\footnotesize for unlabeled data}\end{array}\right.
\intertext{Here, we used $\uk$ as a shorthand notation to directly cover both the labeled and unlabeled case. We can now rewrite this result as softmax function with weighted sums over bottom-up and top-down inputs $\yd$ and $\uk$ as its argument:}
	\dots & = \frac{\exp\!\big(\sum_{\dimd}\yd\log(\Wgencd) + \log(\sum_\classk \Rgenkc \uk)\big)}
	{\sum_{\subc'}\exp\!\big(\sum_{\dimd}\yd\log(\Wgen[\subc'][\dimd])+\log(\sum_\classk \Rgen[\classk][\subc'] \uk)\big)}\nonumber \\
&= \frac{\exp(\Ic)}{\sum_{c^\prime}\exp(\Ic[\subc'][])}\text{, with}\\[6pt]
%
& \Ic = \sum_\dimd \log(\Wgencd)\yd + \log(\sum_\classk \uk \Rgenkc).
\end{align}

\paragraph{M-Step.}
To maximize the free energy with respect to parameters $\Wgencd$ and $\Rgenkc$, we use the method of Lagrange multipliers for constrained optimization:
\begin{align}
 &\frac{\partial \mathcal{F}}{\partial \Wgencd} + \frac{\partial}{\partial \Wgencd} \sum_{\subc'} \lambda_{\subc'}\Big(\sum_{\dimd'} \Wgen[\subc'][\dimd'] - \normA\Big) \overset{!}{=} 0, \label{eq:LagrangeW1} \\
 &\frac{\partial \mathcal{F}}{\partial \Rgenkc} \,+ \frac{\partial}{\partial \Rgenkc}\, \sum_{\classk'}\lambda_{\classk'}\Big(\sum_{\subc'} \Rgen[\classk'][\subc']\, - 1\Big) \overset{!}{=} 0. \label{eq:LagrangeR1}
\end{align}

Starting with the first term of \cref{eq:LagrangeW1} for $\Wgencd$, we insert the free energy \cref{eq:GenFreeEnergy} and evaluate the partial derivative:
\begin{align}
	\frac{\partial}{\partial \Wgencd}\mathcal{F}(\ThetaOld,\Theta) &= \sum_{\inpn,\subc'\!,\classk} p(\subc'\!,\classk|\yVecN,\labelyN,\ThetaOld) \sum_{\dimd'}\frac{\partial}{\partial \Wgencd} \Big(\y^{(\inpn)}_{\dimd'} \log\!\big(\Wgen[\subc'][\dimd']\big) - \Wgen[\subc'][\dimd']\Big) \nonumber \\[-4pt]
	&= \sum_{\inpn} p(\subc|\yVecN,\labelyN,\ThetaOld) \Big(\ydN \frac{1}{\Wgencd} - 1\Big).
\label{eq:OptimizeW1}
\end{align}
The second term of \cref{eq:LagrangeW1}, incorporating the Lagrange multipliers, results in
\begin{equation}
	\frac{\partial}{\partial \Wgencd} \sum_{\subc'}\lambda_{\subc'}\Big(\sum_{\dimd'} \Wgen[\subc'][\dimd'] - \normA\Big) = \sum_{\subc'} \lambda_{\subc'}\sum_{\dimd'} \big( \delta_{\subc \subc'} \delta_{\dimd \dimd'}\big) = \lambda_{\subc}.
\label{eq:OptimizeW2}
\end{equation}
Both terms put back into \cref{eq:LagrangeW1} and multiplied by $\Wgencd$ yields
\begin{align}
	\sum_{\inpn} p(\subc|\yVecN,\labelyN,\ThetaOld) \Big(\ydN - \Wgencd\Big) + \lambda_\subc \Wgencd \overset{!}{=} 0.
	\label{eq:LagrangeW2}
\end{align}
To evaluate the Lagrange multipliers $\lambda_\subc$, we make use of the constraint \cref{eq:Obs} by taking the sum over $\dimd$:
\begin{align}
	\Rightarrow \quad & \sum_{\inpn} p(\subc|\yVecN,\labelyN,\ThetaOld) \Big(\sum_\dimd \ydN - \sum_\dimd \Wgencd \Big) + \lambda_\subc \sum_\dimd \Wgencd	= 0 \nonumber \\[-4pt]
	\Rightarrow \quad & \lambda_\subc = \sum_{\inpn} p(\subc|\yVecN,\labelyN,\ThetaOld) - \frac{1}{\normA}\sum_{\inpn} p(\subc|\yVecN,\labelyN,\ThetaOld)\sum_\dimd \ydN.
	\label{eq:LagrangeWlambda}
\end{align}
Inserting $\lambda_\subc$ back into \cref{eq:LagrangeW2} and canceling opposing terms finally yields the update rule for $\Wgencd$:
\begin{gather}
	\sum_{\inpn} p(\subc|\yVecN,\labelyN,\ThetaOld) \, \ydN - \Wgencd \frac{1}{\normA}\sum_{\inpn} p(\subc|\yVecN,\labelyN,\ThetaOld)\sum_{\dimd'} \y[\!(\inpn)][\dimd'] \overset{!}{=} 0 \\
	\Rightarrow \quad \Wgencd = \normA \frac{\sum_{\inpn} p(\subc|\yVecN,\labelyN,\ThetaOld) \, \ydN}{\sum_{\dimd'} \sum_{\inpn} p(\subc|\yVecN,\labelyN,\ThetaOld) \, \y[\!(\inpn)][\dimd']}.
	\label{eq:MstepWrec}
\end{gather}

The derivation of $\Rkc$ updates follows the same procedure.
Evaluation of the two terms in \cref{eq:LagrangeR1} and multiplication with $\Rgenkc$ gives
\begin{align}
	\sum_{\inpn} p(\subc,\classk|\yVecN,\labelyN,\ThetaOld) + \lambda_\classk \Rgenkc \overset{!}{=} 0.
\label{eq:LagrangeM2}
\end{align}
Using the constraint \cref{eq:PriorC} for $\Rgenkc$, the Lagrange multipliers evaluate to 
\begin{align}
	\lambda_\classk = - \sum_{\inpn,\subc} p(\subc,\classk|\yVecN,\labelyN,\ThetaOld).
	\label{eq:LagrangeMlambda}
\end{align}
Inserting these back into \cref{eq:LagrangeM2}, we arrive at the update rule for $\Rgenkc$:
\begin{align}
	\Rightarrow \quad & \Rgenkc = \frac{\sum_{\inpn}  p(\classk|\subc,\labelyN,\ThetaOld)p(\subc|\yVecN,\labelyN,\ThetaOld)}{\sum_{\subc'} \sum_{\inpn}  p(\classk|\subc',\labelyN,\ThetaOld)p(\subc'|\yVecN,\labelyN,\ThetaOld)}.
	\label{eq:MstepM}
\end{align}

\subsection[Approximate Equivalence of Neural Online Learning]{Approximate Equivalence of Neural Online Learning at Convergence}
\label{SecOnlineLearning}

\newcommand{\al}{\alpha}
\newcommand{\be}{\beta}
\newcommand{\ga}{\gamma}
\newcommand{\la}{\lambda}
\newcommand{\sig}{\sigma}
\newcommand{\sVecPrime}{\vec{s}^{\,\prime}}
\newcommand{\sVecT}{\sVecPrime}

\newcommand{\yVecNPrime}{\vec{y}^{\,(n^{\prime})}}
\newcommand{\yVecCause}{\vec{y}^{\mathrm{\,cause}}}
\newcommand{\yVecTilde}{\vec{\tilde{y}}}
\newcommand{\yVecNTilde}{\vec{\tilde{y}}^{\,(n)}}

\newcommand{\ytN}{\tilde{y}^{(n)}}
\newcommand{\ytC}{\tilde{y}^{(c)}}
\newcommand{\ytVec}{\tilde{\vec{y}}}
\newcommand{\yt}{\tilde{y}}
\newcommand{\yVecBar}{\vec{y}^\mathrm{\,bar}}
\newcommand{\YCal}{{\cal Y}}
\newcommand{\ytD}{\tilde{y}_{d}}
\newcommand{\ytDN}{\tilde{y}_{d}^{(n)}}
\newcommand{\Old}{\mathrm{old}}
\newcommand{\Wid}{W_{id}}
\newcommand{\WidPrime}{W_{id^{\prime}}}
\newcommand{\WidPrimeOld}{W_{id^{\prime}}^{\mathrm{old}}}
\newcommand{\WidOld}{W_{id}^{\mathrm{old}}}
\newcommand{\Whd}{W_{hd}}
\newcommand{\WTilde}{\tilde{W}}
\newcommand{\WEff}{W^{\mathrm{eff}}}
\newcommand{\WBar}{\overline{W}}
\newcommand{\RBar}{\overline{R}}
\newcommand{\WOld}{W^{\prime}}
\newcommand{\Wg}{W^{\mathrm{gen}}}
\newcommand{\WTildeOld}{\tilde{W}^{\mathrm{old}}}
\newcommand{\WW}{{\cal W}}
\newcommand{\WWid}{\WW_{id}}
\newcommand{\WWidPrime}{\WW_{id^{\prime}}}
\newcommand{\WWidPrimeOld}{\WW_{id^{\prime}}^{\prime}}
\newcommand{\WWidOld}{\WW_{id}^{\prime}}
\newcommand{\WWidNew}{\WW_{id}^{\mathrm{new}}}
\newcommand{\WWhd}{\WW_{hd}}
\newcommand{\WWTilde}{\tilde{\WW}}
\newcommand{\WWEff}{\WW^{\mathrm{eff}}}
\newcommand{\WWBar}{\overline{\WW}}
\newcommand{\WcBart}{\overline{W_c}(t)}
\newcommand{\WWOld}{\WW^{\mathrm{old}}}
\newcommand{\WWTildeOld}{\tilde{\WW}^{\mathrm{old}}}
\newcommand{\WDiffThres}{\theta_{\Delta{}W}}
\newcommand{\Scal}{{\cal S}}
\newcommand{\LL}{{\cal L}}
\newcommand{\gen}{\,\mathrm{gen}}
\newcommand{\WNew}{W^\mathrm{new}}
\newcommand{\WNewT}{\tilde{W}^\mathrm{new}}

\newcommand{\Rnn}{R}
\newcommand{\Rnnkc}{\Rnn_{kc}}

\newcommand{\WnncdPrime}{\Wnn_{cd^{\prime}}}
\newcommand{\RnnkcPrime}{\Rnn_{kc^{\prime}}}

\newcommand{\WnncdPrimeOld}{\Wnn_{cd^{\prime}}^{\prime}}

\newcommand{\WnncdOld}{\Wnn_{cd}^{\prime}}
\newcommand{\WnncdNew}{\Wnn_{cd}^{\mathrm{new}}}
\newcommand{\Wnnhd}{\Wnn_{hd}}
\newcommand{\WnnTilde}{\tilde{\Wnn}}
\newcommand{\WnnEff}{\Wnn^{\mathrm{eff}}}
\newcommand{\WnnBar}{\overline{\Wnn}}
\newcommand{\WnnOld}{\Wnn^{\mathrm{old}}}
\newcommand{\WnnTildeOld}{\tilde{\Wnn}^{\mathrm{old}}}
\newcommand{\Wast}{W^{\ast}}
\newcommand{\Ntt}{\tilde{N}}
\newcommand{\zc}[1]{z_c^{(#1)}}

\newcommand{\Fcd}[1]{F_{cd}^{\,(#1)}}
\newcommand{\Gkc}[1]{G_{kc}^{\,(#1)}}
\newcommand{\GkcPrime}[1]{G_{kc'}^{\,(#1)}}

\newcommand{\Fhatcd}[1]{\hat{F}_{cd}^{\,(#1)}}
\newcommand{\FhatcdPrime}[1]{\hat{F}_{c\dPrime}^{\,(#1)}}
\newcommand{\Fhatc}[1]{\hat{F}_{c}^{\,(#1)}}
\newcommand{\FcdBar}{\overline{F}_{cd}}
\newcommand{\FcdPrime}[1]{F_{cd'}^{\,(#1)}}
\newcommand{\FcdPP}[1]{F_{cd''}^{\,(#1)}}
\newcommand{\FcdPrimeBar}{\overline{F}_{cd'}}
\newcommand{\RBig}[1]{R^{\,(#1)}}
\newcommand{\SBig}[1]{S^{\,(#1)}}
\newcommand{\disS}{\displaystyle}%
\newcommand{\kkk}{\hspace{-2.5mm}}
\newcommand{\un}{\vec{u}^{(n)}}

In more detail, to derive \cref{eq:WRLearning}, we first consider the dynamic behavior of the summed weights $\WBar_{c}=\sum_d{}W_{cd}$ and $\RBar_{k}=\sum_c{}R_{kc}$. By taking sums over $d$ and $c$ for \cref{eq:DeltaW,eq:DeltaR} respectively, we obtain
\begin{align}
\Delta{}\WBar_{c}\,=\, \epsilon_{W}\, s_c\,(A\,-\,\WBar_{c})\,, &&
\Delta{}\RBar_{k}\,=\, \epsilon_{R}\, t_k\,(B\,-\,\RBar_{k})\,. \label[pluralequation]{eq:DeltaWRBar}
\end{align}
As we assume $s_c,t_k\geq{}0$, we find that for small learning rates $\epsilon_{W},\epsilon_{R}$ the states $\WBar_{c}=A$ and $\RBar_{k}=B$ are stable (and the only) fixed points of the dynamics for $\WBar_{c}$ and $\RBar_{k}$.
This applies for all $k$ and $c$ and for any $s_c$ and $t_k$ that are non-negative and continuous w.r.t.\ their arguments.

The above result uses an approach developed by \cite{KeckEtAl2012} which we apply here to a hierarchical system with two hidden layers instead of one, and by considering label information.
By assuming normalized weights based on \cref{eq:DeltaWRBar}, we can approximate the effect of iteratively applying \cref{eq:DeltaW,eq:DeltaR} as
\begin{equation} \label{eq:FixedPointsOne}
\Wnn_{cd}^{(n+1)} = A \frac{ \Wnn_{cd}^{(n)}\,+\,\epss_{W}\,s_{c}(\yVecN,\un,\Theta^{(n)})\,\ydN}
{\sum_{\dPrime}\big(\Wnn_{c\dPrime}^{(n)} \,+ \epss_{W}\,s_{c}(\yVecN,\un,\Theta^{(n)})\,\!\y[(\inpn)][\dimd']\,\big)}
\end{equation}
and
\begin{equation} \label{eq:FixedPointsOneR}
R_{kc}^{(n+1)} 
  = B\frac{ R_{kc}^{(n)}\,+ \epss_R\,t_{k}(\sVecN,\un,R^{(n)})\,s_c(\yVecN,\un,\Theta^{(n)})}
         { \sum_{\cPrime}\big(R_{k\cPrime}^{(n)}\,+ \epss_R\,t_{k}(\sVecN,\un,R^{(n)})\,s_{\cPrime}(\yVecN,\un,\Theta^{(n)})\big) }\,,
\end{equation}
where $W^{(n)}$ and $R^{(n)}$ denote the weights at the $n$th iteration of learning, where $\Theta^{(n)}=(W^{(n)},R^{(n)})$, and where $\sVecN=\sVec(\yVecN,\un,\Theta^{(n)})$ to abbreviate notation.
Both equations can be further simplified. Using the abbreviations $\Fcd{n} = s_{c}(\yVecN,\un,\Theta^{(n)})\,\ydN$ and $\Gkc{n} = t_{k}(\sVecN,\un,R^{(n)})\,s_c(\yVecN,\un,\Theta^{(n)})$, we first rewrite \cref{eq:FixedPointsOne,eq:FixedPointsOneR} as
\begin{align}
& W_{cd}^{(n+1)} = A\frac{ \Wnncd^{(n)}\,+ \epss_W\,\Fcd{n} }{\sum_{d'} \big( \WnncdPrime^{(n)} + \epss_W\,\FcdPrime{n}\big)} \quad\, \text{and} \quad\,
R_{kc}^{(n+1)} = B\frac{ \Rnnkc^{(n)}\,+ \epss_R\,\Gkc{n} }{\sum_{d'} \big( \RnnkcPrime^{(n)} + \epss_R\,\GkcPrime{n}\big)}.
\end{align}
Let us suppose that learning has converged after about $T$ iterations.
If we now add another $N$ iterations and repeatedly apply the learning steps, closed-form expressions for the weights $\Wnncd^{(T+N)}$ and $\Rnnkc^{(T+N)}$ are given by
\begin{equation} 
\Wnncd^{(T+N)} = \frac{ \Wnncd^{(T)}\,+ \epss_W\,\sum_{n=1}^N\Fcd{T+N-n} 
                                \prod_{\nPrime=n+1}^N (1 + \frac{\epss_W}{A} \sum_{d'} \FcdPrime{T+N-\nPrime}) }
         { \prod_{\nPrime=1}^N (1 + \frac{\epss_W}{A} \sum_{d'} \FcdPrime{T+N-\nPrime}) }
\label{eq:NNUpdateRepeated}
\end{equation}
and
\begin{equation} 
\Rnnkc^{(T+N)} 
  = \frac{ \Rnnkc^{(T)}\,+ \epss_R\,\sum_{n=1}^N\Gkc{T+N-n} 
                                \prod_{\nPrime=n+1}^N (1 + \frac{\epss_R}{B} \sum_{c'} \GkcPrime{T+N-\nPrime}) }
         { \prod_{\nPrime=1}^N (1 + \frac{\epss_R}{B} \sum_{c'} \GkcPrime{T+N-\nPrime}) }.
 \label{eq:NNUpdateRepeatedR}
\end{equation}

The large products in numerator and denominator of \cref{eq:NNUpdateRepeated,eq:NNUpdateRepeatedR}
can be regarded as polynomials of order $N$ for $\epss_W$ and $\epss_R$, respectively.
Even for small $\epss_W$ and $\epss_R$ it is difficult, however, to argue that higher-order terms of $\epss_W$ and $\epss_R$ can be neglected because of the combinatorial growth of prefactors given by the large products.

We therefore consider the approximations derived for the non-hierarchical model in \cite{KeckEtAl2012}, which were applied to an equation of the same structure as \cref{eq:NNUpdateRepeated,eq:NNUpdateRepeatedR}. At closer inspection of the terms $\Fcd{T+N-n}\!$ and $\Gkc{T+N-n}$, we find that we can apply these approximations also for the hierarchical case. For completeness, we reiterate the main intermediate steps of these approximations below:

Taking \cref{eq:NNUpdateRepeated} as example, we simplify its right-hand-side. The approximations are all assuming a small but finite learning rate $\epss_W$ and a large number of inputs $N$. \Cref{eq:NNUpdateRepeated} is then approximated by 
\begin{align}
\Wnncd^{(T+N)} &\approx \frac{ \Wnncd^{(T)}\,+ \epss_W\,\sum_{n=1}^N \exp\!\big( \frac{\epss_W}{A} (N-n) \sum_{\dPrime}\FhatcdPrime{n}\big)\,\Fcd{T+N-n}}
{\exp\!\big( \frac{\epss_W}{A} N \sum_{\dPrime}\FhatcdPrime{0} \big)}
\label{eq:FirstApprox}\\
 &\approx \exp\!\big( -\frac{\epss_W}{A}\,N\,\sum_{\dPrime}\FhatcdPrime{0} \big)\,\Wnncd^{(T)}
\,+\,\epss_W\,\Fhatcd{0}\,\sum_{n=1}^N \exp\!\big( -\frac{\epss_W}{A}\,n\,\sum_{\dPrime}\FhatcdPrime{0}\big)
\label{eq:SecondApprox}\\
&\approx \Fhatcd{0}\,\frac{ \epss_W\,\exp\!\big( -\frac{\epss_W}{A}\sum_{\dPrime}\FhatcdPrime{0}\big) }{1-\exp\!\big( -\frac{\epss_W}{A}\sum_{\dPrime}\FhatcdPrime{0}\big)}
\,=\,A\,\frac{\Fhatcd{0}}{\sum_{\dPrime}\FhatcdPrime{0}}
\,=\,A\frac{\sum_{n=1}^N\Fcd{T+N-n}}{\sum_{\dPrime}\sum_{n=1}^N\FcdPrime{T+N-n}}\,\label{eq:ThirdApprox}
\end{align}
where $\Fhatcd{n}=\frac{1}{N-n}\sum_{\nPrime=n+1}^N\Fcd{T+N-\nPrime}$ (note that $\Fhatcd{0}$ is the mean of $\Fcd{n}$ over $N$ iterations starting at iteration $T$).

For the first step (\ref{eq:FirstApprox}) we rewrote the products in \cref{eq:NNUpdateRepeated} and used a Taylor expansion \citep[for details, see Supplement of][]{KeckEtAl2012}:
\begin{equation} \label{eq:ProdApprox}
    \prod_{\nPrime=n+1}^N \!\!\!\! (1 + \frac{\epss_W}{A} \sum_{d'} \FcdPrime{T+N-\nPrime})
\approx \exp\!\big( \frac{\epss_W}{A}\,(N-n)\sum_{d'} \FhatcdPrime{n} \big).
\end{equation}

For the second step (\ref{eq:SecondApprox}) we approximated the sum over $n$ in \cref{eq:FirstApprox} by observing that the terms with large $n$ are negligible, and by approximating sums of $\Fcd{T+N-n}$ over $n$ by the mean $\Fhatcd{0}$.
For the last steps, \cref{eq:ThirdApprox}, we used the geometric series and approximated for large $N$
 \citep[for details on these last two approximations, see again Supplement of][]{KeckEtAl2012}.
Furthermore, we used the fact that for small $\epss_W$, $\frac{ \epss_W\,\exp( -\epss_W\,B ) }{1-\exp( -\epss_W\,B )}\approx{}B^{-1}$ (which can be seen, for example, by applying l'H\^opital's rule).

By inserting the definition of $\Fcd{n}$ into (\ref{eq:ThirdApprox}) we finally find:
\begin{equation} \label{AppWFinal}
\Wnncd^{(T+N)} 
  \;\approx\;  
   A\,\frac{ \sum_{n=1}^{N}\,s_{c}(\yVecN,\un,\Theta^{(T+n)})\,\y_d^{\,(n)} }
     {\sum_{\dPrime}\sum_{n=1}^N\,s_{c}(\yVecN,\un,\Theta^{(T+n)})\,\y_{\dPrime}^{\,(n)}}\,.
\end{equation}
Analogously, we find for $R_{kc}$:
\begin{equation} \label{AppRFinal}
\Rnnkc^{(T+N)} 
  \;\approx\;  
   B\,\frac{ \sum_{n=1}^{N}\,t_{k}(\sVecN,\un,\Rnn^{(T+n)})\,s_c^{(n)}}
     {\sum_{\cPrime}\sum_{n=1}^N\,t_{k}(\sVecN,\un,\Rnn^{(T+n)})\, s_{\cPrime}^{(n)}}\,,
\end{equation}
where we again used $\sVecN=\sVec(\yVecN,\un,\Theta^{(T+n)})$ for better readability in the last equation. 
If we now assume convergence, we can replace $\Wnncd^{(T+N)}$ and $\Wnncd^{(T+n)}$ by $\Wnncd$ and $\Rnnkc^{(T+N)}$ and $\Rnnkc^{(T+n)}$ by $\Rnnkc$ to recover \cref{eq:WRLearning} in \cref{sec:NeuralNetwork} with converged weights $\Wnncd$ and $\Rnnkc$.

Note that each approximation is individually very accurate for small $\epss_W$ and large $N$. \Cref{eq:WRLearning} can thus be expected to be satisfied with high accuracy in this case and numerical experiments based on comparisons with EM batch-mode learning verified such high precision.

\section{Computational Details}
\label{AppTrainingDetails}
\subsection{Parallelization on GPUs and CPUs}
\label{AppParallelization}
The online update rules of the neural network \cref{tab:learningrules} are ideally suited for parallelization using GPUs, as they break down to elementary vector or matrix multiplications.
We observed GPU executions with Theano to result in training time speed-ups of over two orders of magnitude compared to single-CPU execution (NVIDIA GeForce GTX TITAN Black GPUs vs. AMD Opteron 6134 CPUs).

Furthermore, we can use the concept of mini-batch training for CPU parallelization or to optimize GPU memory usage. There, the learning effect of a small number $\nu$ of consecutive updates in \cref{eq:DeltaW,eq:DeltaR} is approximated by one parallelized update over $\nu$ independent updates:
\begin{align}
\Delta\!^\nu\WcdN & := \eW \sum_{i=0}^{\nu-1}\Big(\Sc[\subc][(\inpn+i)\,] \y[(\inpn+i)\,][\dimd] - \Sc[\subc][(\inpn+i)\,] \WcdN\Big), & \Wcd^{(\inpn+\nu)\,}  &\approx \WcdN + \Delta\!^\nu\WcdN  \label{eq:DeltaWminibatch}\\
\Delta\!^\nu\RkcN & := \eR \sum_{i=0}^{\nu-1}\Big(\tk[\classk][(\inpn+i)\,] \Sc[\subc][(\inpn+i)\,] - \tk[\classk][(\inpn+i)\,] \RkcN\Big), & \Rkc^{(\inpn+\nu)\,} &\approx \RkcN + \Delta\!^\nu\RkcN. \label{eq:DeltaRminibatch}
\end{align}

The maximal aberration from single-step updates caused by this approximation can be shown to be of $O((\epsilon \nu)^2)$.
Since this effect is negligible for $\epsilon \nu \ll 1$, as experimentally confirmed in \cref{tab:minibatch}, we only consider the mini-batch-size $\nu$ as a parallelization parameter, and not as free parameter that could be chosen to optimize training in anything else than training speed.

\begin{table}[!htb]
  \vspace{6pt}
	\centering
	{
\scriptsize \sffamily \sansmath
\renewcommand{\arraystretch}{1.3}
\newcommand{\midspace}{\hphantom{abc}}
\setlength{\tabcolsep}{3pt}
\begin{tabular}{@{}l ccc@{}}
\toprule
mini-batch size																	& 1											& 10 										& 100 									\\
\midrule
mean test error $[\%]$													& $23.1 \pm 0.2$ 				& $23.2 \pm 0.2$ 				& $23.0 \pm 0.2$ 				\\
std. dev.	$\sigma_\mathrm{err}$	$[\mathrm{pp}]$	& 1.27 									& 1.26 									& 1.24 									\\
\midrule
mean log-likelihood															& $-836.472 \pm 0.005$ 	& $-836.468 \pm 0.005$ 	& $-836.475 \pm 0.005$ 	\\ 
std. dev.	$\sigma_\mathrm{ll}$									& 0.034									& 0.032									&	0.036 								\\
\bottomrule
\end{tabular}
}
	\caption{Results are shown as average over \num{50}~training runs on a small network of $\subC = 30$ hidden units using $\inpN = 3000$ training data points of the MNIST data set. The mini-batch size shows no significant influence neither on the mean nor the variance of the test error or likelihood of the converged solutions.}
	\label{tab:minibatch}
\end{table}

\subsection{Weight Initialization}\label{SecInitDetails}
For the complete setting ($\subC = \classK$), where there is a good amount of labeled data per hidden unit even when labeled data is sparse and the risk of running into early local optima where the classes are not well separated is high, we initialize the weights of the first hidden layer in a modified version of \cite{KeckEtAl2012}:
We compute the mean $m_{\classk\dimd}$ and standard deviation $\sigma_{\classk\dimd}$ of the labeled training data for each class $\classk$ and set $\W[\classk][\dimd] = m_{\classk\dimd} + \mathcal{U}(0,2\sigma_{\classk\dimd})$, where $\mathcal{U}(x_{dn},x_{up})$ denotes the uniform distribution in the range $(x_{dn},x_{up})$. 

For the overcomplete setting ($\subC > \classK$), where there are far less labeled data points than hidden units in the semi-supervised setting, and class separation is no imminent problem, we initialize the weights using all data disregarding the label information.
With the mean $m_d$ and standard deviation $\sigma_d$ over all training data points we set $W_{cd} = m_{d} + \mathcal{U}(0,2\sigma_{d})$.

The weights of the second hidden layer are initialized as $\Rkc = 1/\subC$. The only exception to this rule are the additional experiments on the 20~Newsgroups data set in \cref{sec:20NewsFullyLabeled} for the fully labeled setting. As noted in the text, in this setting we were able to make better use of the recurrent connections of the r-NeSi network and the fully labeled data set by initializing the weights of the second hidden layer as $\Rkc = \delta_{\classk\subc}$.

\subsection{Interactive Labeling with r\textsuperscript{+}-NeSi}\label{sec:InteractiveLabeling}
In \cref{sec:BioInspired}, we trained only the first layer of an ff-NeSi network of 100~hidden units and assigned class labels to the learned representations afterwards by hand.
This already achieved a test error of $(10.53 \pm 0.11)\%$.
To further improve on these results, we can use both the recurrence and the self-labeling of r\textsuperscript{+}-NeSi to our advantage:

First, we can train the second hidden layer also on unlabeled data by setting $\tk = 1/\classK$.
This assumes that all labels are equally likely for all input data points.
This way the network learns the distribution $p(\subc|\R)$ from the input data and can use this information in the recurrent connections.
We can then assign the classes to fields by weighting the labels with this learned distribution $p(\subc|\R) = \frac{1}{\classK} \sum_\classk\Rkc$:
\begin{equation}
  \Rkc =  \frac{\delta_{\labely^{(\subc)}\classk}\sum_{\classk'}\R[\classk'][\subc]^\mathrm{\,old}}{\sum_{\subc'}\delta_{\labely^{(\subc')}\classk}\sum_{\classk'}\R[\classk'][\subc']^\mathrm{\,old}}.
\label{eq:setlabel}
\end{equation}
This way, we do not change the learned information $p(\subc|\R)$ but only set the conditional $p(\classk|\subc,\R)=\frac{\Rkc}{\sum_{\classk'}\R[\classk'][\subc]}$ to $\delta_{\labely^{(\subc)}\classk}$.
Using this distribution for class inference can already significantly decrease the test error, as can be seen by comparison of the `hard max' assignment in \cref{tab:BioResults} to the learned complete distribution (`implicit').

Second, instead of training a network of \num{100}~hidden middle layer units, we can again train a much bigger network of~\num{10000}~hidden units, but still only label 100 of the learned fields.
During further training with self-labeling, the classes of these few labeled fields will also provide class information for the remaining 99\%~of unlabeled fields to learn their associated classes.
While there may be many more informative ways to pick the \num{100}~fields that the supervisor has to label, we here simply chose those fields at random.
Over 10~repetitions, we achieved a test error of $(5.15 \pm 0.26)\%$, which is comparable to the results when training on 100~random labels in the training set.

\section{Detailed Training Results}\label{AppResults}
We performed 100~independent training runs for results obtained on MNIST and 20~Newsgroups in \cref{sec:Results20News,sec:ResultsMNIST}, and 10~independent training runs for the NIST data set in \cref{sec:NIST} with each of the given networks for each label setting with new randomly chosen, class-balanced labels for each training run.
\Cref{tab:rNeSi20News,tab:ffNeSi20News,tab:rNeSiMNIST,tab:ffNeSiMNIST,tab:r+NeSiMNIST,tab:ff+NeSiMNIST,tab:ff+NeSiNISTdigits,tab:r+NeSiNISTdigits,tab:ff+NeSiNISTletters,tab:r+NeSiNISTletters} give a detailed summary of the statistics of the obtained results.
They show the mean test error alongside the standard error of the mean (SEM), the standard deviation (in pp.), as well as the minimal and maximal test error in the given number of runs.
For the networks with self-labeling of unlabeled data (ff\textsuperscript{+}- and r\textsuperscript{+}-NeSi) we only show the semi-supervised settings, as they are identical to their respective standard versions in the fully labeled case.

\begin{table}[t]
  \newcommand\sep{1.3}
  \begin{minipage}{0.495\textwidth}
  	\centering
  	\fontsize{8}{10} \sffamily \sansmath
\renewcommand{\arraystretch}{\sep}
\newcommand{\midspace}{\hphantom{abc}}
\setlength{\tabcolsep}{7pt}
\scriptsize

\begin{tabular}{@{}l r@{}c@{}l c c c}
\toprule
\#labels & \multicolumn{3}{@{}c@{}}{mean test error} & std. dev. & min. & max. \\
\midrule
20 & 70.64 &	$\,\pm\,$	&	0.68 & 6.82 & 55.35 & 88.59 \\
40 & 55.67 &	$\,\pm\,$	&	0.54 & 5.44 & 37.53 & 68.13 \\
200 & 30.59 &	$\,\pm\,$	&	0.22 & 2.22 & 26.97 & 37.57 \\
800 & 28.26 &	$\,\pm\,$	&	0.10 & 1.00 & 26.68 & 31.59 \\
2000 & 27.87 &$\,\pm\,$	&	0.07 & 0.74 & 25.85 & 30.01 \\
11269 & 28.08 &$\,\pm\,$	&	0.08 & 0.78 & 26.29 & 30.25 \\
\bottomrule
\end{tabular}
\setlength{\tabcolsep}{6pt}

  	\vspace{-4pt}
  	\caption{ff-NeSi on 20~Newsgroups.}
  	\label{tab:ffNeSi20News}
  \end{minipage}
  \hfill
  \begin{minipage}{0.495\textwidth}
  	\centering
  	\fontsize{8}{10} \sffamily \sansmath
\renewcommand{\arraystretch}{\sep}
\newcommand{\midspace}{\hphantom{abc}}
\setlength{\tabcolsep}{7pt}
\scriptsize

\begin{tabular}{@{}l r@{}c@{}l c c c}
\toprule
\#labels & \multicolumn{3}{@{}c@{}}{mean test error} & std. dev. & min. & max. \\
\midrule
20 & 68.68 &	$\,\pm\,$	&	0.77 & 7.72 & 49.98 & 85.48 \\
40 & 54.24 &	$\,\pm\,$	&	0.66 & 6.59 & 37.00 & 66.76 \\
200 & 29.28 &	$\,\pm\,$	&	0.21 & 2.09 & 25.90 & 39.60 \\
800 & 27.20 &	$\,\pm\,$	&	0.07 & 0.70 & 25.85 & 29.41 \\
2000 & 27.15 &$\,\pm\,$	&	0.07 & 0.65 & 25.77 & 29.13 \\
11269 & 27.28 &$\,\pm\,$&	0.07 & 0.73 & 26.08 & 29.82 \\
\bottomrule
\end{tabular}
\setlength{\tabcolsep}{6pt}

  	\vspace{-4pt}
  	\caption{r-NeSi on 20~Newsgroups.}
  	\label{tab:rNeSi20News}
  \end{minipage}
  
  \vspace{10pt}
  \begin{minipage}{0.495\textwidth}
   	\centering
   	\fontsize{8}{10} \sffamily \sansmath
\renewcommand{\arraystretch}{\sep}
\newcommand{\midspace}{\hphantom{abc}}
\setlength{\tabcolsep}{7pt}
\scriptsize

\begin{tabular}{@{}l r@{}c@{}l c c c}
\toprule
\#labels & \multicolumn{3}{@{}c@{}}{mean test error} & std. dev. & min. & max. \\
\midrule
10 & 55.46 &	$\,\pm\,$	&	0.57 & 5.72 & 42.49 & 69.62 \\
100 & 19.08 &	$\,\pm\,$	&	0.26 & 2.61 & 13.31 & 24.93 \\
600 & 7.27 &	$\,\pm\,$	&	0.05 & 0.49 & 6.01 & 8.76 \\
1000 & 5.88 &	$\,\pm\,$	&	0.03 & 0.31 & 5.19 & 6.97 \\
3000 & 4.39 &$\,\pm\,$	&	0.02 & 0.15 & 4.01 & 4.89 \\
60000 & 3.27 &	$\,\pm\,$	&	0.01 & 0.08 & 3.08 & 3.46 \\
\bottomrule
\end{tabular}
\setlength{\tabcolsep}{6pt}

    	\vspace{-4pt}
   	\caption{ff-NeSi on MNIST.}
  	\label{tab:ffNeSiMNIST}
  \end{minipage}
  \hfill
  \begin{minipage}{0.495\textwidth}
   	\centering
   	\fontsize{8}{10} \sffamily \sansmath
\renewcommand{\arraystretch}{\sep}
\newcommand{\midspace}{\hphantom{abc}}
\setlength{\tabcolsep}{7pt}
\scriptsize

\begin{tabular}{@{}l r@{}c@{}l c c c}
\toprule
\#labels & \multicolumn{3}{@{}c@{}}{mean test error} & std. dev. & min. & max. \\
\midrule
10 & 29.61 &	$\,\pm\,$	&	0.57 & 5.71 & 20.05 & 46.05 \\
100 & 12.43 &	$\,\pm\,$	&	0.15 & 1.53 & 9.29 & 16.25 \\
600 & 6.94 &	$\,\pm\,$	&	0.05 & 0.49 & 5.72 & 8.44 \\
1000 & 6.07 &	$\,\pm\,$	&	0.03 & 0.28 & 5.24 & 6.78 \\
3000 & 4.68 &$\,\pm\,$	&	0.02 & 0.19 & 4.22 & 5.29 \\
60000 & 2.94 &	$\,\pm\,$	&	0.01 & 0.08 & 2.75 & 3.14 \\
\bottomrule
\end{tabular}
\setlength{\tabcolsep}{6pt}

    	\vspace{-4pt}
   	\caption{r-NeSi on MNIST.}
   	\label{tab:rNeSiMNIST}
  \end{minipage}
  
  \vspace{10pt}
  \begin{minipage}{0.495\textwidth}
   	\centering
   	\fontsize{8}{10} \sffamily \sansmath
\renewcommand{\arraystretch}{\sep}
\newcommand{\midspace}{\hphantom{abc}}
\setlength{\tabcolsep}{7pt}
\scriptsize

\begin{tabular}{@{}l r@{}c@{}l c c c}
\toprule
\#labels & \multicolumn{3}{@{}c@{}}{mean test error} & std. dev. & min. & max. \\
\midrule
10 & 10.91 &	$\,\pm\,$	&	0.86 & 8.64 & 3.96 & 53.15 \\
100 &  4.96 &	$\,\pm\,$	&	0.08 & 0.82 &  3.84 &  9.13 \\
600 & 4.08 &	$\,\pm\,$	&	0.02 & 0.17 & 3.68 & 4.73 \\
1000 & 4.00 &	$\,\pm\,$	&	0.01 & 0.12 & 3.76 & 4.38 \\
3000 & 3.85 &$\,\pm\,$	&	0.01 & 0.11 & 3.64 & 4.14 \\
\bottomrule
\end{tabular}
\setlength{\tabcolsep}{6pt}

    	\vspace{-4pt}
   	\caption{ff\textsuperscript{+}-NeSi on MNIST.}
  	\label{tab:ff+NeSiMNIST}
  \end{minipage}
  \hfill
  \begin{minipage}{0.495\textwidth}
   	\centering
   	\fontsize{8}{10} \sffamily \sansmath
\renewcommand{\arraystretch}{\sep}
\newcommand{\midspace}{\hphantom{abc}}
\setlength{\tabcolsep}{7pt}
\scriptsize

\begin{tabular}{@{}l r@{}c@{}l c c c}
\toprule
\#labels & \multicolumn{3}{@{}c@{}}{mean test error} & std. dev. & min. & max. \\
\midrule
10 & 18.68 &	$\,\pm\,$	&	0.89 & 8.90 &  5.06 & 51.88 \\
100 &  4.93 &	$\,\pm\,$	&	0.05 & 0.49 & 4.26 &  7.32 \\
600 & 4.34 &	$\,\pm\,$	&	0.01 & 0.15 & 3.87 & 4.78 \\
1000 & 4.26 &	$\,\pm\,$	&	0.01 & 0.12 & 3.97 & 4.62 \\
3000 & 4.05 &$\,\pm\,$	&	0.01 & 0.10 & 3.84 & 4.29 \\
\bottomrule
\end{tabular}
\setlength{\tabcolsep}{6pt}

    	\vspace{-4pt}
   	\caption{r\textsuperscript{+}-NeSi on MNIST.}
   	\label{tab:r+NeSiMNIST}
  \end{minipage}
  
  \vspace{10pt}
  \begin{minipage}{0.495\textwidth}
   	\centering
   	\fontsize{8}{10} \sffamily \sansmath
\renewcommand{\arraystretch}{\sep}
\newcommand{\midspace}{\hphantom{abc}}
\setlength{\tabcolsep}{7pt}
\scriptsize

\begin{tabular}{@{}l r@{}c@{}l c c c}
\toprule
\#labels & \multicolumn{3}{@{}c@{}}{mean test error} & std. dev. & min. & max. \\
\midrule
10 & 7.56 &	$\,\pm\,$	&	1.76 & 5.67 & 5.52 & 23.46 \\
100 & 6.20 &	$\,\pm\,$	&	0.16 & 0.51 & 5.49 & 7.08 \\
600 & 6.02 &	$\,\pm\,$	&	0.08 & 0.25 & 5.72 & 6.51 \\
1000 & 6.02 &	$\,\pm\,$	&	0.12 & 0.38 & 5.63 & 6.99 \\
3000 & 5.70 &$\,\pm\,$	&	0.03 & 0.10 & 5.56 & 5.89 \\
344307 & 5.11 &	$\,\pm\,$	&	0.01 & 0.03 & 5.06 & 5.16 \\
\bottomrule
\end{tabular}
\setlength{\tabcolsep}{6pt}

    	\vspace{-4pt}
   	\caption{ff$^+$-NeSi on NIST digits.}
  	\label{tab:ff+NeSiNISTdigits}
  \end{minipage}
  \hfill
  \begin{minipage}{0.495\textwidth}
   	\centering
   	\fontsize{8}{10} \sffamily \sansmath
\renewcommand{\arraystretch}{\sep}
\newcommand{\midspace}{\hphantom{abc}}
\setlength{\tabcolsep}{7pt}
\scriptsize

\begin{tabular}{@{}l r@{}c@{}l c c c}
\toprule
\#labels & \multicolumn{3}{@{}c@{}}{mean test error} & std. dev. & min. & max. \\
\midrule
10 & 9.84 &	$\,\pm\,$	&	2.41 & 7.61 & 5.64 & 34.95 \\
100 & 6.14 &	$\,\pm\,$	&	0.23 & 0.72 & 5.52 & 7.84 \\
600 & 5.83 &	$\,\pm\,$	&	0.14 & 0.45 & 5.43 & 6.50 \\
1000 & 5.94 &	$\,\pm\,$	&	0.12 & 0.39 & 5.46 & 6.49 \\
3000 & 5.72 &$\,\pm\,$	&	0.10 & 0.33 & 5.52 & 6.63 \\
344307 & 4.52 &	$\,\pm\,$	&	0.01 & 0.04 & 4.44 & 4.56 \\
\bottomrule
\end{tabular}
\setlength{\tabcolsep}{6pt}

    	\vspace{-4pt}
   	\caption{r$^+$-NeSi on NIST digits.}
   	\label{tab:r+NeSiNISTdigits}
  \end{minipage}
  
  \vspace{10pt}
  \begin{minipage}{0.495\textwidth}
   	\centering
   	\fontsize{8}{10} \sffamily \sansmath
\renewcommand{\arraystretch}{\sep}
\newcommand{\midspace}{\hphantom{abc}}
\setlength{\tabcolsep}{7pt}
\scriptsize

\begin{tabular}{@{}l r@{}c@{}l c c c}
\toprule
\#labels & \multicolumn{3}{@{}c@{}}{mean test error} & std. dev. & min. & max. \\
\midrule
52 & 55.70 &	$\,\pm\,$	&	0.62 & 1.96 & 52.88 & 58.75 \\
520 & 46.22 &	$\,\pm\,$	&	0.43 & 1.37 & 43.91 & 48.47 \\
3120 & 44.24 &	$\,\pm\,$	&	0.23 & 0.74 & 43.23 & 45.49 \\
5200 & 43.69 &	$\,\pm\,$	&	0.21 & 0.65 & 42.53 & 44.40 \\
15600 & 42.96 &$\,\pm\,$	&	0.28 & 0.88 & 41.55 & 44.38 \\
387361 & 34.66 &	$\,\pm\,$	&	0.05 & 0.15 & 34.45 & 34.86 \\
\bottomrule
\end{tabular}
\setlength{\tabcolsep}{6pt}

    	\vspace{-4pt}
   	\caption{ff$^+$-NeSi on NIST letters.}
  	\label{tab:ff+NeSiNISTletters}
  \end{minipage}
  \hfill
  \begin{minipage}{0.495\textwidth}
   	\centering
   	\fontsize{8}{10} \sffamily \sansmath
\renewcommand{\arraystretch}{\sep}
\newcommand{\midspace}{\hphantom{abc}}
\setlength{\tabcolsep}{7pt}
\scriptsize

\begin{tabular}{@{}l r@{}c@{}l c c c}
\toprule
\#labels & \multicolumn{3}{@{}c@{}}{mean test error} & std. dev. & min. & max. \\
\midrule
52 & 64.97 &	$\,\pm\,$	&	0.85 & 2.70 & 60.88 & 69.71 \\
520 & 54.08 &	$\,\pm\,$	&	0.38 & 1.21 & 51.71 & 55.89 \\
3120 & 43.73 &	$\,\pm\,$	&	0.15 & 0.47 & 42.99 & 44.62 \\
5200 & 41.57 &	$\,\pm\,$	&	0.13 & 0.42 & 40.90 & 42.21 \\
15600 & 37.95 &$\,\pm\,$	&	0.12 & 0.38 & 37.25 & 38.56 \\
387361 & 31.93 &	$\,\pm\,$	&	0.06 & 0.18 & 31.63 & 32.17 \\
\bottomrule
\end{tabular}
\setlength{\tabcolsep}{6pt}

    	\vspace{-4pt}
   	\caption{r$^+$-NeSi on NIST letters.}
   	\label{tab:r+NeSiNISTletters}
  \end{minipage}
\end{table}

\section{Tunable Parameters of the Compared Algorithms}\label{sec:ParameterComparison}
We list in \cref{TabHyperParams} the tunable parameters of each method compared to in \cref{fig:MNIST-sota-Scatter,fig:MNIST-sota-Bar}.
For some of the methods, this estimate only gives a lower bound on the number of tunable parameters, as  parameters of them may have multiple instances, for example, for each added layer in the network.
If a parameter was kept constant for all layers, we only counted it as a single parameter, whereas such parameters that had differing values in different layer were counted as multiple parameters.
An example is the constant number of hidden units in `NN' versus the differing numbers in the layers of the `CNN'.
We also counted such parameters, that were not (explicitly) optimized in the corresponding papers itself, but were taken from other papers (for example, parameters of the ADAM algorithm), or where the reason for the specific choice is not given (like for specific network architectures).

\begin{table}[!b]
	\centering
	\renewcommand{\arraystretch}{1.3}

\begin{tabularx}{\textwidth}{p{2cm}Xr}
\bfseries{\scshape{Method}}  &\bfseries{\scshape{Tunable Hyper-Parameters}} & \bfseries{\scshape{Total}}\\
\hline \\
SVM & $C$ (soft margin parameter) & $1$\\
TSVM		& $C$ (soft margin parameter), $\lambda$ (data-similarity kernel parameter) & $2$\\
NN	 & number of hidden layers (here: 1-15), number of hidden units (same per layer), learning rate  & $3$\\
AGR		& $m,s, \gamma$, dimensionality reduction (for acceleration) & $3-4$\\
NeSi (ours)	& $\subC$ (number of middle layer units), $\normA$ (input normalization), learning rates $\eW$ and $\eR$, $\BvSB$ threshold $\vartheta$ (only for r\textsuperscript{+}- and ff\textsuperscript{+}-NeSi) & $4-5$\\
AtlasRBF	& $\lambda, \gamma, \sigma$, number of neighbors, local manifold dimensionality & $5$\\
Em$^{all}$NN	& NN hyper-parameters, number of layers to embed (here: all), $\lambda$, $m$ (distance parameter) & $6$\\
CNN	& number of CNN layers (here: 6), Patch size, pooling window size (\nth{2} layer), neighborhood radius (\nth{4} layer), \nth{1}, \nth{3}, \nth{5} and \nth{6} layer units, learning rate & $\geq 9$\\
M1+M2		& M1: number of hidden layers (here: 2), number of hidden units per layer, number of samples from posterior, M2:  number of hidden layers (here: 1), number of hidden units, $\alpha$, RMSProp: learning rate, first and second momenta & $\geq 10$ \\ 
DBN-rNCA        & number of layers (here: 4),  number of hidden units per layer; RBM learning rate, momentum, weight-decay, RBM epochs, NCA epochs, $\lambda$ (tradeoff parameter), & $\geq 11$\\
EmCNN		& CNN hyper-parameters, number of layers to embed, $\lambda$, $m$ (distance parameter), & $\geq 12$ \\
VAT & number of layers (here: 2 - 4), number of hidden units per layer, $\lambda$, $\epsilon$, $I_p$, ADAM \citep{KingmaBa2015}: learning rate $\alpha$, $\epsilon_{\mathrm{ADAM}}$, exponential decay rates $\beta_1$ and $\beta_2$; batch normalization \citep{IoffeSzegedy2015}: mini-batch size for labeled and mixed set & $\geq 12$\\
Ladder & number of hidden layers (here: 5), number of hidden units per layer, noise level~$\mathbf n^{(l)}$, denoising cost multipliers $\lambda^{(l)}$ for each layer, ADAM \citep{KingmaBa2015}: learning rate $\alpha$, $\epsilon_{\mathrm{ADAM}}$, iterations until annealing phase, linear decay rate; batch normalization \citep{IoffeSzegedy2015}: minibatch size & $\geq 18$\\
\end{tabularx}

	\caption{Tunable hyperparameters of the algorithms compared on the MNIST data set.}
	\label{TabHyperParams}
\end{table}


\clearpage
\fancyhead[LE,RO]{}
{
\setlength{\bibsep}{2.5pt plus 0.3ex}
\small
\bibliography{bibliography}
}

\end{document}